\documentclass[10pt,twocolumn,letterpaper]{article}

\usepackage[pagenumbers]{wacv} 

\usepackage[utf8]{inputenc} 
\usepackage[T1]{fontenc}    
\usepackage{url}            
\usepackage{booktabs}       
\usepackage{amsfonts}       
\usepackage{nicefrac}       
\usepackage{microtype}      
\usepackage{tablefootnote}
\usepackage{longtable}

\usepackage{graphicx}
\usepackage{amsmath}
\usepackage{amssymb}
\usepackage{pifont}
\usepackage{makecell}
\usepackage{multirow}
\usepackage{algpseudocode}
\usepackage{algorithm2e}
\RestyleAlgo{ruled}
\usepackage{bbm}
\usepackage{chapterbib}
\usepackage[dvipsnames]{xcolor} 
\newcommand{\cmark}{\ding{51}}%
\newcommand{\xmark}{\ding{55}}%

\usepackage[pagebackref,breaklinks,colorlinks]{hyperref}

\usepackage[capitalize]{cleveref}
\crefname{section}{Sec.}{Secs.}
\Crefname{section}{Section}{Sections}
\Crefname{table}{Table}{Tables}
\crefname{table}{Tab.}{Tabs.}

\def\wacvPaperID{2431} 
\def\confName{WACV}
\def\confYear{2025}

\begin{document}

\title{DLCR: A Generative Data Expansion Framework via Diffusion for Clothes-Changing Person Re-ID}

\author{Nyle Siddiqui\\
Center for Research in Computer Vision\\
University of Central Florida\\
{\tt\small nyle.siddiqui@ucf.edu}
\and
Florinel Alin Croitoru\\
University of Bucharest, Romania\\
{\tt\small alincroitoru97@gmail.com}
\and
Gaurav Kumar Nayak\\
Indian Institute of Technology, Roorkee\\
{\tt\small gauravkumar.nayak@mfs.iitr.ac.in}
\and
Radu Tudor Ionescu\\
University of Bucharest, Romania\\
{\tt\small raducu.ionescu@gmail.com}
\and
Mubarak Shah\\
Center for Research in Computer Vision\\
University of Central Florida\\
{\tt\small shah@crcv.edu}
}
\maketitle

\begin{abstract}
With the recent exhibited strength of generative diffusion models, an open research question is if images generated by these models can be used to learn better visual representations. While this generative data expansion may suffice for easier visual tasks, we explore its efficacy on a more difficult discriminative task: clothes-changing person re-identification (CC-ReID). CC-ReID aims to match people appearing in non-overlapping cameras, even when they change their clothes across cameras. Not only that current CC-ReID models are constrained by the limited diversity of clothing in current CC-ReID datasets, but generating additional data that retains important personal features for accurate identification is a current challenge. To address this issue we propose DLCR, a novel data expansion framework that leverages pretrained diffusion and large language models (LLMs) to accurately generate diverse images of individuals in varied attire. We generate additional data for five benchmark CC-ReID datasets (PRCC, CCVID, LaST, VC-Clothes, and LTCC) and \textbf{increase their clothing diversity by \boldmath{$10$}$\times$, totaling over \boldmath{$2.1$}M generated images}. DLCR employs diffusion-based text-guided inpainting, conditioned on clothing prompts constructed using LLMs, to generate synthetic data that only modifies a subject's clothes, while preserving their personally identifiable features. With this massive increase in data, we introduce two novel strategies -- progressive learning and test-time prediction refinement -- that reduce training time and boost CC-ReID performance. We validate our method through extensive ablations and experiments, showing massive improvements when training previous CC-ReID methods on our generated data. On the PRCC dataset, we obtain a large top-1 accuracy improvement of $11.3\%$  by training CAL, a state-of-the-art (SOTA) method, with DLCR-generated data. We publicly release our code and generated data for each dataset here: \url{https://github.com/CroitoruAlin/dlcr}.
\end{abstract}

\section{Introduction}
\label{sec:intro}
In computer vision research, improved or novel model architectures for discriminative tasks are commonly proposed and shown to learn better visual representations than their predecessors. Discriminative methods operate under the assumption that their training datasets are sufficiently large and diverse, with the limiting factor being the model's ability to learn the desired objective. However, when there is a lack in diversity in the commonly used datasets for some task, the limiting factor in this case becomes the training data itself. Collecting additional in-domain data can be both difficult and expensive, and generative models have only recently become viable enough to generate additional, high-fidelity training data (which we refer to as \textit{generative data expansion}). Generative data expansion has been previously explored for supervised or self-supervised learning on easier discriminative tasks, such as ImageNet classification \cite{azizi2023synthetic,tian2024stablerep}. While there are other unexplored, more difficult discriminative visual tasks that may benefit even more from generative data expansion, the difficulty lies in generating accurate, in-domain data. In this work, we tackle this problem in the context of CC-ReID, which is difficult in nature and, due to its relative infancy, a majority of datasets suffer from only containing a small number ($2$-$5$) of clothing outfits per subject \cite{Wan-CVPR-2020,PRCC,Qian-ACCV-2020,Gu-CVPR-2022,shu2021large}.

In Person Re-identification \cite{Ye-PAMI-2022,Zahra-PR-2023}, the primary objective is to match an individual across images or videos captured by cameras at different times and locations. This task becomes notably challenging due to disparities in the body poses, lighting, backgrounds, viewpoints, and potential occlusions present across cameras. 
Traditional Re-ID approaches \cite{Chen-ICCV-2019,Dai-ICCV-2019,Park-AAAI-2020}, also known as \textit{short-term Re-ID}, assume that individuals are observed for relatively brief durations and that their attire remains constant throughout videos. However, these assumptions often fall short of practical real-world scenarios. For instance, people's clothes can change when captured across different camera feeds at different times and days. 
Consequently, a new branch of Re-ID, known as \textit{clothes-changing Re-ID} \cite{Qian-ACCV-2020,Gu-CVPR-2022,Jin-CVPR-2022,Han-CVPR-2023}, has recently emerged to reliably identify people even when their clothing constantly changes over extended periods of time. Recent CC-ReID works commonly make discriminative modifications to traditional Re-ID model architectures and loss formulations to disentangle structural (clothes-independent) and appearance information (clothes-dependent) in the learned feature space. However, these discriminative approaches are inherently constrained by the aforementioned low clothing variations present in current CC-ReID datasets.


\begin{figure*}[ht!]
    \centering
    \resizebox{\linewidth}{!}{
    \includegraphics{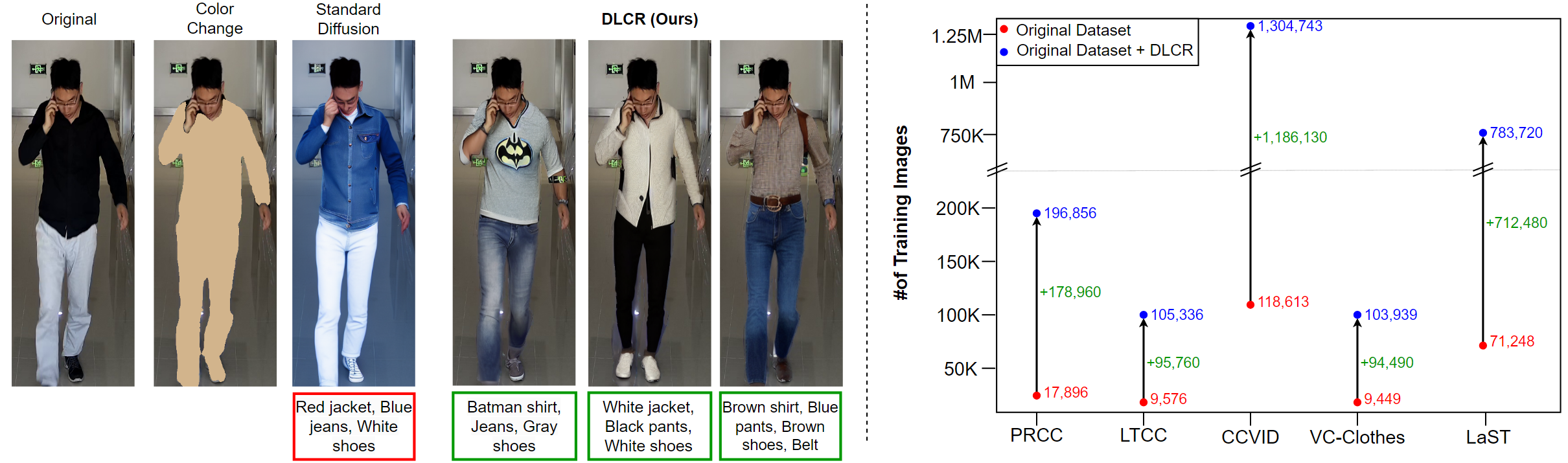}
    }
    \caption{Best viewed with zoom. \textbf{(Left)} Comparing alternative data expansion approaches with our method. Simply changing the color of the clothes using our clothing masks preserves the subject's ID information, but lacks diversity. Using standard diffusion without inpainting can \textit{sometimes} introduce diversity, but the ID-related information in an image is destroyed. Our method strongly increases clothing diversity, even with specific and difficult prompts, and preserves the subject's ID-related information for effective CC-ReID. \textbf{(Right)} DLCR significantly increases the size of training data for five benchmark CC-ReID datasets, generating over $2.1$M diverse, clothes-changed images.}
    \label{fig:motivation}
\end{figure*}

To address these issues, in this work we propose an effective data expansion framework for CC-ReID (named \emph{DLCR}) that increases the number of clothing variations in CC-ReID datasets by intelligently harnessing generative AI and foundation models (\textbf{D}iffusion and large \textbf{L}anguage models for \textbf{C}C-\textbf{R}eID). DLCR consists of two stages: (1) Data Generation and (2) Re-ID Training and Prediction Refinement. In the first stage, our objective is to enrich a given CC-ReID training dataset by generating additional samples where clothing items are artificially changed through inpainting. With more diverse training images of a particular person in different clothes, a CC-ReID model will learn better clothes-invariant person features during training. To achieve this objective, we utilize diffusion models \cite{Ho-NeurIPS-2020, rombach2022high,croitoru2023diffusion,Nichol-ICML-2021}, which are capable of generating realistic images and inpainting for image editing. However, diffusion models often struggle with preserving the intricate details of a person's identity during image generation (see Fig.~\ref{fig:motivation}). We overcome this by leveraging a human parsing method \cite{Li-TPAMI-2020} to produce binary masks that mark only the clothing regions in an image. We then use this binary mask to retain ID-specific portions in an image during diffusion inpainting, such as the face, hair, and body shape, thereby only augmenting the subject's clothes. Furthermore, existing CC-ReID datasets only use scalar values for each unique clothes ID and lack textual descriptions of specific clothing items, preventing the use of text-guided diffusion models for data generation. Thus, to overcome this issue, we generate explicit clothing descriptions by jointly leveraging a large visual-language model (\textit{LLaVA} \cite{Liu-NeurIPS-2023}) and a large language model (\textit{LLaMA} \cite{Touvron-Arxiv-2023}). 


In the second stage of DLCR, we propose two effective strategies in which our generated data can be sophisticatedly utilized during CC-ReID training and testing. Firstly, we train various CC-ReID models using our generated data from stage 1 and show how our enriched training data improves performance in clothes-changing settings. To accommodate both the large increase in data and the higher clothing variety of the generated samples, we employ a \textit{progressive learning} strategy during training, where we gradually introduce the generated clothing variations of a person. This allows for training on $10\times$ more data without drastically increasing training time, as well as further assists the model in learning better clothes-invariant person features. During inference, we apply our \textit{prediction refinement} strategy which is specifically tailored for a CC-ReID retrieval task. We create different variations of a given query by artificially changing the subject's attire, then ensemble their retrieved similarity scores from the gallery to obtain a refined final prediction. 

We summarize our contributions as follows:

\begin{itemize}
\item DLCR is the first to implement a text-guided diffusion approach, in conjunction with foundational language models, 
to synthesize multiple images of a person with different clothes in a CC-ReID dataset (Sec.~\ref{sec:stage1}). 
\item We propose a progressive learning strategy during training that judiciously combines our generated data with the original training set, improving performance and heavily reducing training time (Sec.~\ref{sec:Progressive}). 
\item We employ a novel prediction refinement strategy at test-time, where variants of a given query image are created by changing the subject's clothes, and their retrieval scores are ensembled to further boost performance (Sec.~\ref{sec:Refinement}). 
\item We demonstrate the effectiveness of our method on four benchmark CC-ReID datasets (Sec.~\ref{sec:state-of-the-art}). Our method also yields a significant performance boost when existing approaches (both standard Re-ID and CC-ReID) utilize our generated data during training (Sec.~\ref{sec:improving_existing}). 
\item \textbf{We publicly release our code and generated CC-ReID data, totaling over \boldmath
{$2.1$}M generated images, for future use.}
\end{itemize}

\section{Related Works}
\label{sec:related}
 In standard Re-ID, clothing is considered a static attribute, which leads standard Re-ID models to erroneously couple clothing items with specific subjects. CC-ReID is a more difficult paradigm that requires models to intrinsically learn person-specific features that are robust to covariate information, especially clothing items. To achieve this robustness, both discriminative \cite{hermans2017defense,guo2023semantic,Hong-CVPR-2021,PRCC, li2020gait,eom2019learning,zhang2019gait, Liu-ICCV-2023,wu2017depth,haque2016recurrent} and generative \cite{ma2017pose,wei2018person,ge2018fd, liu2018pose,zheng2019joint,Xu-IJCAI-2021} approaches for CC-ReID have been explored. Discriminative approaches focus on elaborate model architectures \cite{li2020learning,Gu-CVPR-2022} and loss formulations \cite{hermans2017defense,guo2023semantic,Hong-CVPR-2021}, while generative methods aim to synthesize data that increases the quality or diversity of an existing training set. The common goal of these methods is to enforce the learning of clothes-invariant person features, thereby improving CC-ReID performance on datasets in which subjects are captured in multiple different clothing outfits.

\noindent\textbf{Discriminative methods:} A standard approach to learning clothes-invariant features is to use some form of identification loss paired with a metric learning loss, \ie~contrastive or triplet loss \cite{hermans2017defense,guo2023semantic,Hong-CVPR-2021,PRCC}. Alternatively, some methods specifically aim to disentangle the appearance information from the person-relevant information in RGB images \cite{li2020gait,eom2019learning,zhang2019gait} for CC-ReID. Tangential to metric learning, other works have also explored the use of adversarial learning for CC-ReID \cite{li2020learning,Gu-CVPR-2022}. 
Discriminative methods are not limited to loss formulations, as novel architectural designs have also been proposed to accommodate additional clothes-agnostic cues, such as face extraction \cite{Arkushin-WACV-2024} and 3D body shape reconstruction \cite{Liu-ICCV-2023}. Similarly, auxiliary input modalities to RGB, such as human pose \cite{wang2022pose,su2017pose}, depth \cite{wu2017depth,haque2016recurrent}, and gait \cite{Jin-CVPR-2022,li2023depth} have been proposed as multi-modal discriminative approaches to CC-ReID. Separate from input modalities, other works directly augment the learned feature space for better semantic control \cite{Han-CVPR-2023} or robustness to covariate information \cite{song2019generalizable}. However, these discriminative methods are constrained by the inherent limitations of the datasets they are trained on. For example, an insufficient number of clothing outfits in a dataset can be detrimental to metric learning-based CC-ReID models \cite{Han-CVPR-2023}. Thus, our work proposes a generative method that directly improves the clothing diversity in a dataset, better equipping any downstream CC-ReID model to learn clothes-invariant features. 

\noindent
 \textbf{Generative methods:} 
With the aforementioned lack of such diverse datasets, the previously dominant approach has been to use Generative Adversarial Networks (GANs) \cite{goodfellow2014generative} to synthesize additional training data for specific learning objectives, such as pose-invariance \cite{ma2017pose,wei2018person,ge2018fd}, clothes-invariance \cite{liu2018pose,zheng2019joint,Xu-IJCAI-2021}, and improving general performance \cite{zheng2017unlabeled,li2018adversarial,huang2018multi}. However, these GAN-based works also suffer from constraints such as limited diversity/mode collapse \cite{zheng2019joint,liu2019identity} and the limited capability of only transferring existing styles from one subject to another \cite{deng2018image,zhong2018generalizing,zhong2019invariance}. The most important aspect is that GANs are dataset-specific, meaning they require separate training on each dataset before data generation. Furthermore, one of the most successful GAN-based works \cite{zheng2019joint} does not directly use its generated data during training, but rather transfers its dataset-specific generator encoder to the discriminator to produce better clothes-invariant person features, reducing generalizability. On the other hand, diffusion models \cite{Nichol-ICML-2021} have recently overtaken GANs in a multitude of other vision domains \cite{diff-beats-gans,croitoru2023diffusion,saharia-neurips-22,saharia2022image} as a superior class of generative models, yet have not been applied to the CC-ReID domain. Furthermore, the previously discussed limitations of GANs can be overcome by large-scale, pretrained diffusion models which have shown impressive zero-shot generation abilities on a multitude of out-of-domain tasks \cite{ramesh2021zero,rombach2022high,nichol2022glide}.
DLCR is the first generative approach to harness the benefits of diffusion models for CC-ReID, and consequently, uses the same generator for every dataset. In a novel fashion, we leverage a text-conditioned diffusion inpainting model to synthesize completely new clothing outfits onto any subject in a dataset, separating the process between data generation and discriminative CC-ReID training.

\section{Proposed Approach}

\label{sec:methodology}
DLCR is a generic data expansion method that can be applied to any CC-ReID dataset, wherein the given training data is enriched to improve data diversity for better learning of clothes-invariant features. Our method contains two stages: Data Generation (Sec.~\ref{generation}) and Re-ID Training and Prediction Refinement (Sec.~\ref{discriminative-training}). In stage 1, DLCR utilizes semantic human masks, a visual language model (LLaVA), a large language model (LLaMA), and text-conditioned diffusion inpainting 
to controllably change a subject's clothes in an image.
In stage 2, we leverage our generated data, in conjunction with our progressive learning strategy, to train CC-ReID models with $10\times$ more data with negligible impacts to training time. We also propose a test-time prediction refinement strategy specifically tailored for CC-ReID retrieval. 
Before explaining the method in detail, we provide a brief background on diffusion models.

\noindent\textbf{Diffusion models:} are a class of likelihood-based generative models that learn a data distribution by parameterizing a Markov chain through the use of forward and backward processes. Given a data sample, $x_0$ 
$\sim q(x_0)$, the forward process of diffusion iteratively adds Gaussian noise, $\epsilon$, to $x_0$ for $T$ timesteps until $x_T$ is almost 
a sample from a standard Gaussian distribution. Then, the backward process leverages a neural network to learn the reverse distribution $q(x_{t-1} | x_t)$. Thus, a novel data point from the original data distribution, $q(x_0)$, can be simply generated 
by sampling $x_T$ from $\mathcal{N}(\textbf{0}, \textbf{I})$ and applying the Markov reverse steps until $x_0$ is reached. 

\noindent\textbf{Forward Process:} At each timestep, Gaussian noise is added with a timestep dependent variance $\beta_t$ to obtain $q(x_t | x_{t-1}) = \mathcal{N}(x_t;\sqrt{1 - \beta_t}x_{t-1}, \beta_t\textbf{I})$. We can also efficiently sample $x_t$ in a single step \cite{Ho-NeurIPS-2020} as follows: 
\begin{align}
\label{eq1}
    x_t \sim q(x_t | x_{0}) = \mathcal{N}(x_t;\sqrt{ \Bar{\alpha}_t}x_{0}, (1-\Bar{\alpha_t})\textbf{I}),
\end{align}
where $\alpha_{t} = 1- \beta_t$ and $\Bar{\alpha}_t = \prod_{i=0}^t \alpha_i$.
\\
\noindent\textbf{Reverse Process:} The reverse distribution  $q(x_{t-1} | x_t)$ is approximated with a parameterized model $p_\theta$. With small enough chosen values of $\beta_t$, $q(x_{t-1} | x_t)$ will be Gaussian, so a neural network with parameters $\theta$ can be trained to learn the mean $\boldsymbol{\mu}_\theta(x_t, t)$ and variance $\mathbf{\Sigma}_\theta(x_t, t)$ at each timestep $t$:
\begin{equation}
    p_\theta(x_{t-1} | x_t) = \mathcal{N}(x_{t-1}; \boldsymbol{\mu}_\theta(x_t, t), \mathbf{\Sigma}_\theta(x_t,t)).
\end{equation}
Ho \etal~\cite{Ho-NeurIPS-2020} simplifies this objective by expressing 
$\boldsymbol{\mu}_\theta(x_t, t) = \frac{1}{\sqrt{\alpha_t}}\left(\left\lVert x_t - \frac{1-\alpha_t}{\sqrt{1-\Bar{\alpha}_{t}}}\epsilon_\theta(x_t,t)\right\rVert\right),$
and propose to learn to predict $\epsilon_\theta(x_t,t)$ using the loss function $\mathcal{L}_{simple} = E_{t,x_t,\epsilon}(\lVert\epsilon-\epsilon_\theta(x_t,t)\rVert^2)$
, where $\mathbf{\Sigma}_\theta(x_t,t)$ is fixed to $\beta_t\mathbf{I}$. Our method uses text-conditioned diffusion, where a text prompt is used to guide the generation at every timestep, starting from timestep $t=T$ to $t=0$. The reverse process can simply be modified to incorporate a condition $y$ as follows:
\begin{align}
\label{eq_sampling_diffusio}
    p_\theta(x_{t-1} | x_t, y) = \mathcal{N}(x_{t-1}; \boldsymbol{\mu}_\theta(x_t, t, y), \beta_t \mathbf{I}).
    \vspace{-0.15in}
\end{align}

\subsection{Stage 1: Data Generation}
\label{sec:stage1}
\begin{figure*}
    \centering
    \includegraphics[width=\linewidth]{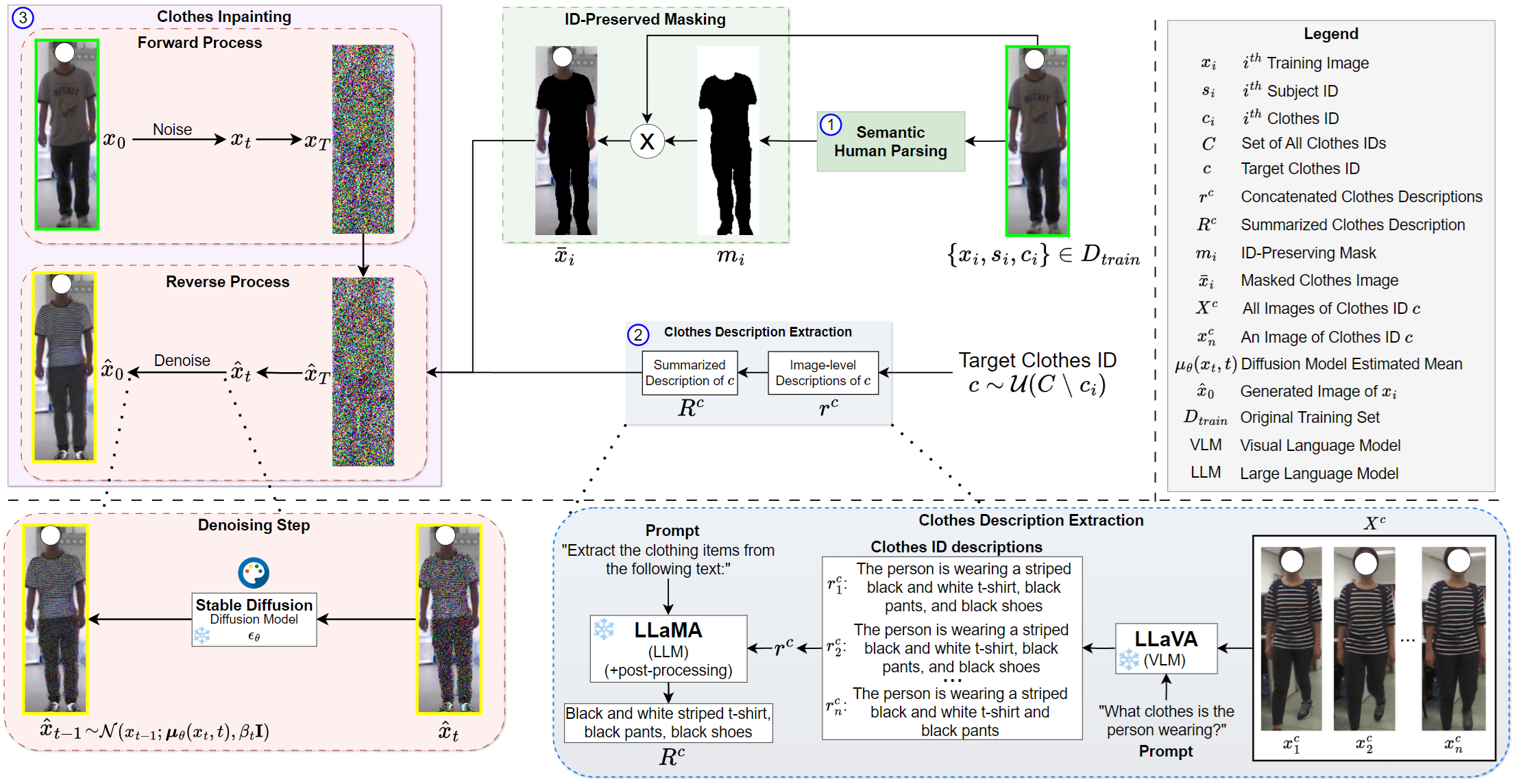}
    \caption{Overview of DLCR data generation (stage 1). Given a training image $x_i$, we apply an ID-preserving mask, $m_i$, where only the clothing regions are marked for inpainting. A target clothing ID, $c$, is randomly selected from the training dataset and a textual description of the clothing items belonging to $c$ is extracted. Specifically, we use LLaVA to extract frame-wise descriptions for every image in $X^c$, and use LLaMA to obtain a summarized description $R^c$. This description is used as a prompt for diffusion inpainting to artificially change a subject's clothing. Our ID-preserving diffusion inpainting model 
    synthesizes quality generated data $\hat{x}_0$ for improved downstream CC-ReID training. \vspace{-0.4cm}
    }
    \label{fig:model-figure}
\end{figure*}

In CC-ReID, the training dataset $D_{train}=\{(x_i, s_i, c_i)\}_{i=1}^{M}$ is a collection of $M$ samples, where the $i^{th}$ sample is a triplet consisting of an RGB image $x_i$, subject (person) ID $s_i$, and clothes ID $c_i$. Generating images of a particular subject, $s_i$, in different clothing outfits can enrich the training data by improving its diversity. Specifically, in this stage, we generate multiple variants of a training image, $x_i$, by artificially inpainting the clothes of subject $s_i$. We achieve this objective by using a text-conditioned diffusion inpainting model (Stable Diffusion Inpainting \cite{rombach2022high}), where text prompts describe the outfits to be generated in the clothing regions specified by a human-parsed semantic mask. 
These components (also shown in Fig.~\ref{fig:model-figure}) are described below. 

\label{generation}

\noindent
\textbf{Semantic Mask \& Text Extraction:}
\label{mask_llm}
While generating different images of a subject, it is important to preserve the person-relevant information. However, diffusion models struggle to retain this information due to the varied and complex nature of human faces and body shapes. Thus, we apply a semantic human parsing method \cite{Li-TPAMI-2020} to each image, $x_i$, to obtain an ID-preserving binary mask, $m_i$, where only the upper-clothes, lower-clothes, and footwear regions are marked for inpainting. This process is computationally negligible, with throughput as high as 60 masks per second. This allows the person ($s_i$) features, such as the face, hair, and general body structure, to be preserved in the generated samples, even after clothes inpainting---an especially critical aspect for enriching CC-ReID data. In general, the set of images belonging to a clothes ID, $c$, can be represented by $X^c=\{x^c_i\}_{i=1}^n$. After applying ID-preserving semantic masks to these images, we obtain a set of masked images $\bar{X}^c=\{x^c_{i} \circ m_{i}\}_{i=1}^n$.
While CC-ReID datasets provide clothes IDs, they are usually simple 
scalar values that do not contain descriptive information as to what specific clothing items are present in an image.
To extract these additional clothing descriptions for each clothes ID 
in $D_{train}$, we utilize LLaVA \cite{Liu-NeurIPS-2023} and LLaMA \cite{Touvron-Arxiv-2023} (see Fig.~\ref{fig:model-figure}). A naive approach would be to simply generate descriptions of clothing items using an LLM, e.g. LLaMA, or to create random clothing descriptions. While this is likely to increase the diversity of the dataset, and consequently, the generalization capacity of the downstream CC-ReID models, it does not alleviate dataset-specific biases. We use LLaVA in order to obtain descriptions of clothing items that are present in the dataset, aiming to reduce the dataset-specific bias (see supplementary). This forces the CC-ReID models to focus on identity features, and ignore clothing features. 

First, we prompt LLaVA to obtain the information of the clothing items for the  upper and lower body, as well as the footwear a person is wearing in each image $x_i^c \in X^c$. We denote the set of text responses obtained from LLaVA as $r^c=\{r^c_i\}_{i=1}^n$, where $r^c_i$ is the clothes description of image $x_i^c$. However, the subject’s visibility can vary across images of a particular clothes ID, $c$, due to changes in occlusions, lighting, or viewpoints. This can lead LLaVA to occasionally respond with missing or incorrect clothing items on certain images of $X^c$. To mitigate this issue, we pass the image-based responses, $r^c$, as input to LLaMA and prompt it to summarize them into a single clothing description. Consequently, we obtain a holistic clothes description, $R^c$, for a particular clothes ID $c$. 
Through this summarization, LLaMA helps to produce accurate clothing descriptions and overcomes the issue of missing clothing items.  We further post-process $R^c$ if there are multiple clothing descriptions obtained from LLaMA for a particular body part. We keep the most frequently occurring descriptions across the images of $c$. We \textbf{do not train/fine-tune} the LLMs as we only use them in an inference setting with an average throughput of roughly 4 samples per second. 
For more details regarding our use of LLMs and their impact on the final results, please refer to the supplementary. 
 
\noindent
\textbf{Diffusion-Based Clothes Inpainting:} \label{inpainting_section}
A text-conditioned diffusion inpainting model, paired with our extracted clothing descriptions and semantic masks, can be used to controllably generate diverse, synthetic samples. To this end, we propose, for the first time, the use of a diffusion inpainting model for CC-ReID. We first construct a set $C=\{c_i\}_{i=1}^L$ containing all the unique clothes IDs in the training data $D_{train}$. For each training image $x_i$ with subject ID $s_i$ and clothes ID $c_i$, we uniformly sample $K$ different clothes IDs, $c_{j} \sim \mathcal{U}(C), \forall j \in \{1,..., K\}$ where $j \neq i$. For a sampled clothes ID $c_{j}$, we obtain its holistic clothes description $R^{c_{j}}$ by leveraging LLaVA and LLaMA (Sec.~\ref{mask_llm}). For a given image $x_i$, we get the ID-preserved masked image $\bar{x}_i$ by applying its corresponding human-parsed semantic mask, $m_i$. To generate $K$ diverse images of the same subject (person $s_i$) corresponding to the $K$ sampled clothes IDs, we utilize the extracted clothes descriptions $\{R^{c_{j}}\}_{j=1}^{K}$ and masked image $\bar{x}_i$ as inputs to a pretrained Stable Diffusion model \cite{rombach2022high} for ID-preserving inpainting. The masked image $\bar{x}_i$ is used in the forward diffusion process (Eq.~\eqref{eq1}), to  preserve ID-related information. The reverse denoising process (Eq.~\eqref{eq_sampling_diffusio}) is conditioned on the prompt $R^{c_{j}}$ to generate the clothes-changed images $\{\hat{x}_{ij}\}_{j=1}^K$ of subject $s_i$ in $x_i$, as follows:
\begin{align}
\label{eq_inpainting}
    \hat{x}_{ij} \sim p_\theta({x}_0 | \bar{x}_i, 
    R^{c_{j}}), \forall j \in \{1,...,K\}.
\end{align}

We \textbf{do not train/fine-tune} the pretrained diffusion model, but solely use it in an inference setting, thus keeping computational requirements low with a throughput of roughly one generated image per second. Thus, we construct a set of generated data $G_{train} = \{(\hat{x}_{ij}, s_i, c_{ij})_{j=1}^K\}_{i=1}^{M}$ for each aforementioned CC-ReID dataset, which can be directly utilized to enrich the original training data set, $D_{train}$. It is this data, containing over $2.1$M generated images in total across multiple CC-ReID datasets, that we release for public use, as summarized in Table \ref{tab:dataset_stats}. 

\begin{table}[t]
    \centering
    \caption{Dataset statistics for each CC-ReID dataset with and without DLCR. (*) We do not publicly release our generated data for LTCC due to its licensing restrictions.}

        \begin{tabular}{c|c|c}
            \Xhline{3\arrayrulewidth}

         \multirow{2}{*}{Dataset} & \multicolumn{2}{c}{Training Images} \\
         \cline{2-3}
         & Original  & Generated \\
         
         \Xhline{2\arrayrulewidth}
         PRCC & $17,896$ & $178,960$ \\
         \hline
         LTCC* & $9,576$* & $95,760$*  \\
        \hline
         CCVID & $118,613$ & $1,186,130$  \\
        \hline
         VC-Clothes & $9,449$ & $94,490$ \\
         \hline
         LaST & $71,248$ & $712,480$  \\
         \Xhline{3\arrayrulewidth}
         Total & - & \boldmath{$2,267,820$}
    \end{tabular}

    \label{tab:dataset_stats}
\end{table}

\subsection{Stage 2: Re-ID Training and Refinement}
\label{discriminative-training}
Our generated training set $G_{train}$ can be used to enrich the original training data $D_{train}$, improving the discriminative capabilities of any Re-ID model. One simple way to utilize $G_{train}$ is to concatenate it with ${D_{train}}$ into one dataset, then use this merged dataset to train a Re-ID model. However, the model struggles to effectively utilize $G_{train}$ when all the clothes-changed variations are introduced at once during training (discussed in supplementary). Moreover, the training time of the CC-ReID models would be heavily impacted with the generated training data being $10\times$ larger in size. 
In the supplementary, we provide a t-SNE visualization of real vs.~generated images showing that the generated data exhibits a higher variance. 
To gently and more efficiently accommodate CC-ReID models to the diverse distribution of the generated data, we propose a progressive learning approach 
that sophisticatedly increases the clothing diversity of a subject as training progresses. Furthermore, the novel ability to controllably inpaint realistic clothes using our method can be extended even further than just during training. Specifically, we propose a prediction refinement strategy 
that uses our diffusion-based inpainting method as a unique test-time augmentation to provide a more holistic and ensemble-based decision during evaluation.
\noindent
\subsubsection{Progressive Learning Strategy}
\label{sec:Progressive}
To mitigate impacts to training time, we gradually introduce $G_{train}$ 
by injecting our generated data into each training batch in a pairwise fashion. During batch construction, we pair every original training image in the batch with one of its corresponding generated samples, effectively doubling the original batch size. For brevity, consider one image from the original training set $x
\in D_{train}$ and its corresponding generated samples $P^x = \{\hat{x}_{i}\}_{i=1}^K$. A generated image from $P^x$ can be chosen as a pair for $x$ through random sampling: $\hat{x} \sim \mathcal{U}(P^x)$. Due to the large increase in clothing diversity and number of images in $G_{train}$, we show that incrementally introducing the clothes-changed generated images from $P^x$ as training progresses yields better results. Moreover, since we generate roughly $10\times$ more training data for each CC-ReID dataset, we found that our progressive learning strategy enables models to effectively utilize our generated data without major impacts on training time (see supplementary). 
Specifically, we perform progressive learning by first dividing $P^x$ into four distinct, equal partitions $P^x = P_1 + P_2 + P_3 + P_4$. Then, during the early stages of training, only the generated images $\hat{x} \in P_1$ are used for sampling when finding a pair for $x$, where $\hat{x} \sim \mathcal{U}(P_1)$. As training continues, we 
periodically increase the sampling range for the generated images by incorporating more partitions. For instance, $x$ is paired with a randomly sampled generated image $\hat{x} \sim \mathcal{U}(\{P_1 + P_2\})$ after $N$ epochs of training have been completed. After $2N$ completed training epochs, the generated pair image is sampled by $\hat{x} \sim \mathcal{U}(\{P_1 + P_2 + P_3\})$ until eventually, every generated image in $P^x$ is 
used for sampling during batch construction after $3N$ training epochs: $\hat{x} \sim \mathcal{U}(P^x)$. In this manner, we increase the difficulty of learning clothes-invariant features by controlling the number of clothes-changed generated images seen during the early stages of training. Therefore, the model gradually learns to ignore more clothing items as training progresses due to the increasing diversity of subject attire.

\subsubsection{Prediction Refinement Strategy}
\label{sec:Refinement}

While we have only discussed using DLCR for generating CC-ReID training data, we hereby explain that we can also leverage DLCR at test-time by employing our inpainting method (Sec. \ref{inpainting_section}) to augment a given query image by changing its clothes. For a given query, we create $l$ 
augmented query variants. 
For each query variant, we then retrieve the top-$m$ similarity scores from the gallery and ensemble them to mitigate clothing bias when making a prediction. Consequently, our ensembling method yields a refined similarity score estimate of a subject ID, resulting in more accurate top-1 retrievals. 
Algorithm~\ref{refinement_algo} describes the 
detailed steps of our ensembling process. 
Our algorithm takes a query image, the $l$ augmented versions of the query, the gallery, and the CC-ReID model as input. 
The first step computes the features of the query ($f^q$), augmented query variants $\{\hat{f}^q_i\}_{i=1}^l$ and gallery images $\{\hat{f}^g_i\}_{i=1}^n$ (Line $1$). In step two, for the original query, we retrieve the top-$m$ most similar examples, $I^m$, from the gallery (Lines $2$-$3$). 
In the fourth step, each unique subject $\{s_i \in S\}_{i=1}^m$ in $I^m$ is assigned an initial score. We repeat this retrieval process for each augmented query variant $\{\hat{f}^q_k\}_{k=1}^l$, resulting in their own top-$m$ retrievals from the gallery, $\{I_k^m\}_{k=1}^l$ (Line $5$). For each $I_k^m$, we update the similarity scores only for the subjects present in $S$ (Line $6$). At the end, we obtain the set of subject IDs $S$, with their corresponding refined similarity scores. Hence, for a given query image, we leverage our diffusion-based inpainting and ensembling algorithm to yield a better clothes-invariant prediction. 
Algorithm \ref{refinement_algo} is also extremely computationally efficient, with the entire process having a throughput of $\sim640$ images/sec with a memory constraint of $<2$ GB.

\SetNlSty{textt}{}{.}
\SetCommentSty{mycommfont}
\begin{algorithm}[!t]
\scriptsize{
\caption{\small{Refinement of Re-ID Predictions\label{alg_sampling_ddpm}}}
\label{refinement_algo}
\SetAlgoLined
\SetKwInOut{Input}{Input}  
\textbf{Input}: $x^q$: query image, $\{(x^g_i, s_i)\}_{i=1}^n$: labeled gallery images, 
\\$\{\hat{x}^q_i\}_{i=1}^l$: $l$ augmented variants of $x^q$, $m$: size of query-gallery retrieval, 
\\$r_\psi$: CC-ReID model, $\cos(\cdot, \cdot)$: cosine similarity.\\
\textbf{Output}: Set $S$ of subject IDs. \\

{\color{blue}{\tcc{1: Extract 
features}}}
\nl $f^q \gets r_\psi(x^q); f^g_i \gets r_\psi(x^g_i), \forall i \in \{1,...,n\};$  $\hat{f}^q_i \gets r_\psi(\hat{x}^q_i), \forall i \in \{1,...,l\}$ \\
{\color{blue}{\tcc{2: Get top-m similar gallery samples}}}
\nl $I \gets \text{argsort}_{i \in \{1,...,n\}}\left\{\cos(f^q, f^{g}_{i})
\right\}$ \hspace{-0.02in}{\color{blue}{\tcp{\hspace{0.01in}desc order}}}
\nl $I^m \gets I\left[:m\right]$\\
{\color{blue}{\tcc{3: Assign similarity scores to 
subject id 
in top-m}}}
\nl $S \gets \{s_i | i \in I^m\}$; $score\left[ s \right] 
\gets \sum_{j \in I^m}
{ \mathbbm{1}_{\{s\}}(s_j)\frac{\cos(f^q, f^{g}_{j})}{\max\limits_{k \in I^m} \cos(f^q, f^{g}_{k})}},\forall s \in S$\\  
\For{$k = 1, \dots, l$}{

\nl $I_k \gets \text{argsort}_{i \in \{1,...,n\}}\left\{cos(\hat{f}^q_k, f^{g}_{i})
\right\}$; $I_k^m \gets I_k\left[:m\right]$;
$S_k \gets \left\{s_i| i \in I_k^m\right\}$ \\
{\color{blue}{\tcc{4: Update the similarity scores using the inpainted examples}}}
    \For{$s \in S_k$}{ 
        \If{$s \in S$}{
\nl            $score\left[s\right] \gets score\left[s\right] + \sum_{j \in I_k^m}{\mathbbm{1}_{\{s\}}(s_j) \frac{\cos(\hat{f}^q_k, f^{g}_{j})}{\max\limits_{p \in I_k^m} \cos(\hat{f}^q_k, f^{g}_{p})}}$\\
        }
    }
}
\nl \textbf{return} $S$ 
}
\end{algorithm}

\section{Experiments}
\label{sec:experiments}

\noindent\textbf{Implementation details:}
We use CAL \cite{Gu-CVPR-2022} as the baseline CC-ReID model in our experiments as it generally performed the best with our generated data across all datasets. We follow all of their training hyper-parameters unless specified otherwise, but note that our method is compatible with any Re-ID model as discussed in Sec.~\ref{sec:add-on}. We use Stable Diffusion \cite{rombach2022high} as our diffusion inpainting model. We set the number of generated clothing outfits to $K=10$. In progressive learning, we increase the sampling range after every $N=5$ epochs. For our prediction refinement strategy, we set the number of augmented queries $l=6$ and keep the top $m=5$ retrievals. We resize input images to $384 \times 192$, and train downstream Re-ID models with an original batch size of $32$ which is doubled to $64$ when paired with generated data, however note that this has a negligible effect on Re-ID training time. All experiments are conducted on a single NVIDIA A100 80GB GPU, but only the data generation stage requires at least one GPU with $\le$ 48GB of memory, whereas training and deploying the CC-ReID models can be done on smaller consumer GPUs ($\le$12GB). Additional throughput and computational complexity details are provided in the supplementary.

\subsection{Comparison with State-of-the-Art Methods}
\label{sec:state-of-the-art}
We report results on the clothes-changing settings of four recent CC-ReID datasets: PRCC \cite{PRCC}, LTCC \cite{Qian-ACCV-2020}, VC-Clothes \cite{Wan-CVPR-2020}, 
and CCVID \cite{Gu-CVPR-2022}. Following the standard protocol described in each dataset, we use the top-1 accuracy and mean average precision (mAP) as evaluation metrics, and use the official dataset query and gallery splits. 
We apply stage 1 (Sec.~\ref{generation}) and stage 2 (Sec.~\ref{discriminative-training}) of DLCR to CAL \cite{Gu-CVPR-2022} to obtain our results shown in Table \ref{tab:main_results}. With the help of DLCR-generated data, we achieve new SOTA on 3 out of 4 datasets with CAL, with the largest SOTA improvement of $5.3\%$ in top-1 accuracy on PRCC. Note that while CCFA \cite{Han-CVPR-2023} is the only published model to outperform DLCR on LTCC, their source code is not available, hence we were not able to train CCFA using our generated data. With available code, we hypothesize that DLCR-generated data would most likely improve CCFA and achieve SOTA on LTCC.

\begin{table}[ht!]
    \small
    \caption{Comparisons with previous CC-ReID models on the PRCC, LTCC, CCVID, and VC-Clothes datasets in the clothes-changed (CC) setting. DLCR essentially outperforms every model across four datasets, with the green text denoting improvements over the CAL baseline. $\dagger$ denotes reproduced results using provided open-source code.}
    \centering

    \resizebox{\columnwidth}{!}{
    \begin{tabular}{c|c|c|c|c}
         \Xhline{3\arrayrulewidth}
         Dataset & Model & Venue & Top-1 & mAP  \\
        \Xhline{2\arrayrulewidth}
         \multirow{9}{*}{PRCC} 
         & GI-ReID \cite{Jin-CVPR-2022} & CVPR '22 & 37.5 & - \\
         & AFD-Net \cite{Xu-IJCAI-2021} & IJCAI '21 & 42.8 & - \\
         & RCSANet \cite{Huang-ICCV-2021} & ICCV '21 & 50.2 & 48.6 \\
         & 3DSL \cite{Chen-CVPR-2021} & CVPR '21 & 51.3 & - \\
         & FSAM \cite{Hong-CVPR-2021} & CVPR '21 & 54.5 & - \\
         & CAL (baseline) $\dagger$ \cite{Gu-CVPR-2022} & CVPR '22 & 55.2 & 55.8 \\
         & AIM \cite{AIM} & CVPR '23 & 57.9 & 58.3 \\
         & CCFA$^{1}$ \cite{Han-CVPR-2023} & CVPR '23 & 61.2 & 58.4\\\cline{2-5}
         
         & Ours & - & \textbf{66.5 (\textcolor{Green}{+11.3})} & \textbf{63.0 (\textcolor{Green}{+7.2})} \\
        \Xhline{3\arrayrulewidth}
         \multirow{3}{*}{CCVID} 
        & CAL (baseline) \cite{Gu-CVPR-2022} & CVPR '22 & 81.7 & 79.6 \\
        & 3DInvarReID \cite{Liu-ICCV-2023} & ICCV '23 & 84.3 & 81.3 \\\cline{2-5}
        & Ours & - & \textbf{88.0 (\textcolor{Green}{+6.3})} & \textbf{84.5 (\textcolor{Green}{+4.9})}\\
        
        \Xhline{3\arrayrulewidth}
        \multirow{5}{*}{VC-Clothes}
        & GI-ReID \cite{Jin-CVPR-2022} & CVPR '22 & 64.5 & 57.8 \\
        & 
        3DSL \cite{Chen-CVPR-2021} & CVPR '21 & 79.9 & 81.2 \\
        & FSAM \cite{Hong-CVPR-2021} & CVPR '21 & 78.6 &  78.9 \\

        & CAL (baseline) \cite{Gu-CVPR-2022} & CVPR '22 & 85.8 & 79.8 \\\cline{2-5}
        & Ours & - & \textbf{87.1 (\textcolor{Green}{+1.3}}) & \textbf{81.1 (\textcolor{Green}{+1.3}}) \\
       
       \Xhline{3\arrayrulewidth}
        \multirow{7}{*}{LTCC} 
         & GI-ReID \cite{Jin-CVPR-2022} & CVPR '22 & 26.7 & 12.7 \\
         & 3DSL \cite{Chen-CVPR-2021} & CVPR '21 & 31.2 & 14.8 \\
         & FSAM \cite{Hong-CVPR-2021} & CVPR '21 & 38.5 & 16.2 \\
         & CAL (baseline) $\dagger$ \cite{Gu-CVPR-2022} & CVPR '22 & 39.5 & 18.0 \\
         & AIM \cite{AIM} & CVPR '23 & 40.6 & 19.1 \\
         & CCFA\tablefootnote{Source code not publicly available to train on our generated data} \cite{Han-CVPR-2023} & CVPR '23 & \textbf{45.3} & \textbf{22.1}\\\cline{2-5}

         & Ours & - & 41.3 (\textcolor{Green}{+1.8}) & 19.6 (\textcolor{Green}{+1.6}) \\
       \Xhline{3\arrayrulewidth}
    \end{tabular}
    }

    \label{tab:main_results}
\end{table}

\subsection{Improving Existing Methods using DLCR}
\label{sec:improving_existing}

Our results in Table \ref{tab:main_results} are achieved by applying DLCR to CAL since it generally performed the best with our data across all datasets. While both stage 1 (Sec.~\ref{generation}) and stage 2 (Sec.~\ref{discriminative-training}) of DLCR can be used with any Re-ID model, we show in Table \ref{tab:add-on} that simply adding our stage 1 generated data during training with any existing method provides significant CC-ReID performance gains. We train three standard Re-ID models (PCB \cite{Sun-ECCV-2018}, MGN \cite{Wang-ICM-2018}, and HPM \cite{Fu-AAAI-2018}) and three CC-ReID models (CAL, AIM \cite{AIM}, and GEFF \cite{Arkushin-WACV-2024}) with our added generated data on the clothes-changing protocol of PRCC. Even when applied to standard Re-ID models that are not specifically tailored to be clothes-agnostic, our generated data boosts CC-ReID top-1 accuracy by nearly $12-30\%$ and mAP by $12-22\%$. Moreover, training CC-ReID models, which are already tailored for clothes-invariance, with our additional generated data provides further top-1 and mAP improvements of $2-8\%$. In the supplementary, we show ablations on each proposed component of DLCR and further improvements when both stages of DLCR are applied.

\begin{table}[t!]
     \caption{Results on the clothes-changing protocol of PRCC when training various standard Re-ID and CC-ReID models using only stage 1 of DLCR. Our generated data improves performance for any model. $\dagger$ denotes reproduced results using open-source code.}
    \centering
    \small
    \setlength\tabcolsep{0.5em}
    \resizebox{0.85\columnwidth}{!}{
    \begin{tabular}{c|c|c|c}
        \Xhline{3\arrayrulewidth}
         Re-ID Type & Model & Top-1 & mAP  \\
        \Xhline{2\arrayrulewidth}
         \multirow{6}{*}{Standard} & PCB \cite{Sun-ECCV-2018}  & 41.8 & 38.7\\
         & PCB + DLCR  & \textbf{53.3 (\textcolor{Green}{+11.5})} & \textbf{50.7 (\textcolor{Green}{+12.0})}\\ 
         \cline{2-4}
         & MGN \cite{Wang-ICM-2018}  & 33.8 & 35.9\\
         & MGN + DLCR  & \textbf{62.5 (\textcolor{Green}{+28.7})} & \textbf{57.6 (\textcolor{Green}{+21.7})} \\ 
         \cline{2-4}
         & HPM \cite{Fu-AAAI-2018} & 40.4 & 37.2 \\
         & HPM + DLCR  & \textbf{56.0 (\textcolor{Green}{+15.6})} & \textbf{50.9 (\textcolor{Green}{+13.7})}\\
        \Xhline{3\arrayrulewidth}
         & CAL $\dagger$ \cite{Gu-CVPR-2022}  & 55.2 & 55.8 \\
         & CAL + DLCR  & \textbf{62.9 (\textcolor{Green}{+7.7})} & \textbf{60.9 (\textcolor{Green}{+5.1})}\\
         \cline{2-4}
         Clothes & AIM $\dagger$ \cite{AIM} & 55.7 & 56.3 \\
         Changing & AIM + DLCR & \textbf{60.2 (\textcolor{Green}{+4.5})} & \textbf{59.0 (\textcolor{Green}{+2.7})} \\ 
         \cline{2-4}
         & GEFF \cite{Arkushin-WACV-2024} & 83.6 & 64.0 \\
         & GEFF + DLCR &\textbf{84.6 (\textcolor{Green}{+1.0})} & \textbf{66.0 (\textcolor{Green}{+2.0})}\\ 
         \Xhline{3\arrayrulewidth}
    
    \end{tabular}
    }
    
    \label{tab:add-on}
\end{table}
\label{sec:add-on}

\subsection{Performance on Out-of-Distribution Data}
\label{cross-dataset}

We show in this section that CAL+DLCR improves performance on out-of-distribution testing data as well. 
As previously mentioned, CAL introduces architectural changes and loss formulations, such as a clothes-classifier and clothes adversarial loss, to enforce clothes-agnostic learning. However, discriminative changes like these may only enable a model to perform better on data that it was trained on, while poorly generalizing when evaluated on unseen data. To this point, we show results in Table \ref{tab:cross-dataset} where we take a pretrained CAL model and a pretrained CAL+DLCR model on the LTCC dataset, and evaluate them in a zero-shot fashion on other CC-ReID datasets (PRCC, LaST \cite{shu2021large}, and VC-Clothes). Top-1 accuracy gains ranging from $4\%$ to $8\%$ are seen across all three CC-ReID datasets, with similar gains on mAP, ranging from $\approx 1\%$ to $6\%$. 
Therefore, DLCR both enhances a model's performance on a single dataset and even on out-of-distribution testing data.

\begin{table}[ht!]
    \small
    \centering
    \caption{Comparisons between CAL+DLCR and CAL models pretrained on LTCC when evaluated on other datasets. CAL+DLCR consistently beats CAL when evaluated on other datasets in a zero-shot fashion, exhibiting DLCR's discriminative benefits extend beyond just the training dataset.
    }
    
    \begin{tabular}{c|c|c|c|c}
        \Xhline{3\arrayrulewidth}
        \multirow{2}{*}{Model}   & \multicolumn{2}{c|}{Dataset} & \multirow{2}{*}{Top-1} &  \multirow{2}{*}{mAP} \\
        \cline{2-3}
        & Training& Evaluation & & \\
        \Xhline{2\arrayrulewidth}

        \multirow{3}{*}{CAL~\cite{Gu-CVPR-2022}} &\multirow{6}{*}{LTCC} & PRCC & 37.5 & 35.4  \\
        & & LaST  & 20.7 & 3.5 \\
        & & VC-Clothes  & 5.3 & 5.6 \\
        \cline{1-1}
        \cline{3-5}
        \multirow{3}{*}{CAL+DLCR} & & PRCC & \textbf{45.5} & \textbf{38.8}  \\
        & & LaST & \textbf{25.0} & \textbf{4.2} \\
        & & VC-Clothes & \textbf{11.8} & \textbf{11.9}  \\
        
         \Xhline{3\arrayrulewidth}

    \end{tabular}
    \label{tab:cross-dataset}
\end{table}

\section{Conclusion}

\label{sec:conclusion}
In this work, we proposed the first use of diffusion for generative data expansion in CC-ReID. Through diffusion inpainting, DLCR is capable of controllably generating completely new clothing items onto a subject, while retaining person-specific information. Guided by clothing prompts extracted using foundational language models, DLCR synthesizes additional training data to increase the clothing diversity in a CC-ReID dataset. During CC-ReID training, a progressive learning strategy is applied to efficiently utilize this generated data. To better refine a model's prediction during inference, we employ a test-time augmentation. We demonstrate that DLCR significantly improves CC-ReID performance and achieves SOTA results across multiple benchmark datasets. Finally, we release all of our generated data for the benchmark datasets used in this paper for future research. 

{\small
\bibliographystyle{ieee_fullname}
\bibliography{egbib}
}

\clearpage 

\twocolumn[{%
 \begin{center}
  \Large\textbf{DLCR: A Generative Data Expansion Framework via Diffusion for Clothes-Changing Person Re-ID - Supplementary Material}\\
\end{center}
}]


\renewcommand{\thetable}{A\arabic{table}}
\renewcommand{\thefigure}{A\arabic{figure}}
\setcounter{table}{0}

\section{Overview}
\label{sec:overview}
We organize the supplementary material into the following sections: 
\begin{itemize}
    \item Licensing information regarding datasets used in this paper, as well as any dataset/privacy/ethics information regarding our data, is detailed in Section \ref{sec:data-info}.
    \item Section \ref{sec:ex+abl} contains ablations for each component of DLCR.
    \item Section~\ref{add-qual-results} contains additional qualitative examples of our publicly released clothes-changed generated data.
    \item Sections \ref{add-on-stage2} and \ref{visualizations} provide more quantitative and qualitative CC-ReID results to exhibit the superiority of DLCR over previous works.
    \item Section \ref{sec:disc-guidance} details an optional discriminator that can be added to the DLCR data generation stage to generate synthetic data closer to the original training data distribution, marginally improving results.
    \item Section \ref{sec:llm-motif} motivates our specific use of LLaVA and LLaMA.
    \item Section \ref{sec:prog-learning} provides more insights and motivation behind our progressive learning method.
    \item Section \ref{sec:clothing-ablation} provides an ablation on the number of generated images per training image in DLCR.
    \item Section \ref{sec:time} details the throughput and space complexity of each DLCR component.
    \item Lastly, we briefly detail some possible limitations and future avenues of work in Section \ref{sec:limitations}.

\end{itemize}

\section{Licensing/Dataset Information}
\label{sec:data-info}
In this section, we provide information regarding licensing, since we are modifying and publicly releasing data that originates from previous CC-ReID datasets. 

\noindent\textbf{Dataset Licenses:} Most importantly to note, the \textbf{LTCC} dataset license explicitly states that no modification or redistribution of the dataset is allowed unless by Fudan University. Thus, while we show that DLCR does work on LTCC and improves results, we will not publicly release our generated data for LTCC. Researchers are free to use DLCR to re-generate data on their own for the LTCC dataset if they have requested and been granted acccess to the dataset. The \textbf{PRCC} dataset does not follow a standardized license, but simply states that the dataset may only be used for academic purposes, which our work falls under. The \textbf{CCVID} dataset explicitly states their dataset falls under a CC BYNC-SA 4.0 license, which allows for sharing, modifying and/or adapting the data in anyway as long as credit is given, the data is not used for commercial purposes. Our generated data falls under the same license, and we do not impose any additional limitations. 
The \textbf{VC-Clothes} dataset follows the Apache License 2.0, which also allows for redistribution and modification for academic purposes. In summary, DLCR follows all licensing requirements for every dataset in the paper, with the only note being not publicly releasing our generated LTCC data. 

\noindent\textbf{Release, Maintenance, and Ethical Use of DLCR-generated Data:} As mentioned throughout the paper, we publicly release our generated data with full accessibility (no PI contact required, full data and code available) at this URL: \url{https://huggingface.co/datasets/ihaveamoose/DLCR}.
Since we release our data on a publicly available data storage website, there is no maintenance requirements or future access restriction for our generated data. Regarding ethical use of our data, since we only modify the clothing items of human subjects in pre-existing CC-ReID, there are no additional privacy or ethical concerns that are not already addressed by these datasets when they were released. However, we do cover the face when possible in the paper to further protect privacy (we do show the face occasionally in order to exhibit certain qualities of DLCR-generated data). Regarding the license of our data, we choose CC BYNC-SA 4.0 only because CCVID requires our data to be released under that license due to our use and modification of their data. \textbf{For all intents and purposes, we allow for full and unrestricted academic use of our code and data as long as we are properly credited in the work.}

\section{Additional Experiments and Ablations}
\label{sec:ex+abl}
\subsection{Ablations}
\label{sec:ablation}
To demonstrate the utility of each proposed component of DLCR, 
we perform ablations on the PRCC dataset and show these results in Table \ref{tab:ablations}. The first row contains results obtained with a baseline CAL model without using DLCR. 

\noindent{\textbf{Effectiveness of ID-preserving generated data:}} Simply adding our generated data during Re-ID training (Table \ref{tab:ablations}, row $3$) leads to the most significant improvement in performance, with a $5.5\%$ increase in top-1 accuracy and $3\%$ increase in mAP over the baseline (Table \ref{tab:ablations}, row $1$). In row $2$ of Table \ref{tab:ablations}, where data is generated using standard image-to-image diffusion, we see marginal improvements as opposed to our method, which uses ID-preserving masks and inpainting. This shows that increasing the variety of clothes-changing training samples, while still preserving the subject's ID-related information, is integral to improving CC-ReID performance.

\noindent{\textbf{Effectiveness of LLM prompts:}} Row $3$ in Table \ref{tab:ablations} corresponds to generating data by using random clothes prompts as a text condition. In row $4$ of Table \ref{tab:ablations}, we show the impact of using LLMs to extract the textual clothing descriptions in a dataset for text conditioning, with a $2\%$ boost in performance. More information regarding the use and impact of LLMs in DLCR can be found in Sec. \ref{sec:llm-motif}. 


\noindent{\textbf{Effectiveness of progressive learning:}}
Since DLCR generates multiple clothes-changed samples for each training image, 
$G_{train}$ is larger in size than 
$D_{train}$. To effectively utilize $G_{train}$ during training, while also mitigating additional training time, we gradually introduce new clothing variations at the mini-batch level with our progressive learning strategy. This further increases model performance by another $2\%$ (Table \ref{tab:ablations}, row 5), highlighting the importance of elaborate procedures when training with generated data.  

\noindent{\textbf{Effectiveness of prediction refinement:}} As discussed in Sec. $3.2.2$, our diffusion-based inpainting method can also be used as a query augmentation at test-time. The model's predictions on these augmentations are ensembled using Alg.~$1$ to obtain refined similarity scores for each subject, resulting in better test-time predictions and yielding a further $1.5\%$ improvement (Table \ref{tab:ablations}, row 6).

\begin{table}[tb!]
    \caption{Ablations on each proposed component of DLCR on the PRCC dataset. The addition of each component yields consistent improvements in performance. Baseline CAL results are given in the first row. Cumulative performance gains of each component with respect to the baseline are shown in green.}
    \centering
    \resizebox{\linewidth}{!}{
    \begin{tabular}{c|c|c|c|c|c}
    \Xhline{3\arrayrulewidth}
         \makecell{Generated \\ Data} & LLMs & \makecell{Progressive \\ Learning} & \makecell{Prediction \\Refinement} & Top-1 & mAP \\
         \hline
         \textcolor{red}{\xmark} & \textcolor{red}{\xmark} &  \textcolor{red}{\xmark} & \textcolor{red}{\xmark} & 55.2 & 55.8 \\
         \hline
         \textcolor{Green}{\cmark} (Standard Diffusion) & \textcolor{red}{\xmark}  & \textcolor{red}{\xmark} & \textcolor{red}{\xmark} & 55.7 \textbf{\textcolor{Green}{+0.5}} & 55.9 \textbf{\textcolor{Green}{+0.1}} \\
         \hline
          \textcolor{Green}{\cmark} (Ours) & \textcolor{red}{\xmark}&  \textcolor{red}{\xmark} & \textcolor{red}{\xmark} & 60.7 \textbf{\textcolor{Green}{+5.5}} & 58.9 \textbf{\textcolor{Green}{+3.1}} \\
         \hline
         \textcolor{Green}{\cmark} (Ours) & \textcolor{Green}{\cmark} &  \textcolor{red}{\xmark} & \textcolor{red}{\xmark} & 62.9 \textbf{\textcolor{Green}{+7.7}} & 60.9 \textbf{\textcolor{Green}{+5.1}} \\
         \hline
         \textcolor{Green}{\cmark} (Ours) & \textcolor{Green}{\cmark}  & \textcolor{Green}{\cmark} & \textcolor{red}{\xmark} & 65.0 \textbf{\textcolor{Green}{+9.8}} & 62.4 \textbf{\textcolor{Green}{+6.6}} \\
         \hline
         \textcolor{Green}{\cmark} (Ours) & \textcolor{Green}{\cmark} & \textcolor{Green}{\cmark} & \textcolor{Green}{\cmark} & 66.5 \textbf{\textcolor{Green}{+11.3}} & 63.0 \textbf{\textcolor{Green}{+7.2}} \\
         \Xhline{3\arrayrulewidth}
         
    \end{tabular}
    }
    \label{tab:ablations}
\end{table}



\section{Additional Generated Examples}
\begin{figure*}[ht!]
    \centering
    \includegraphics[width=\linewidth]{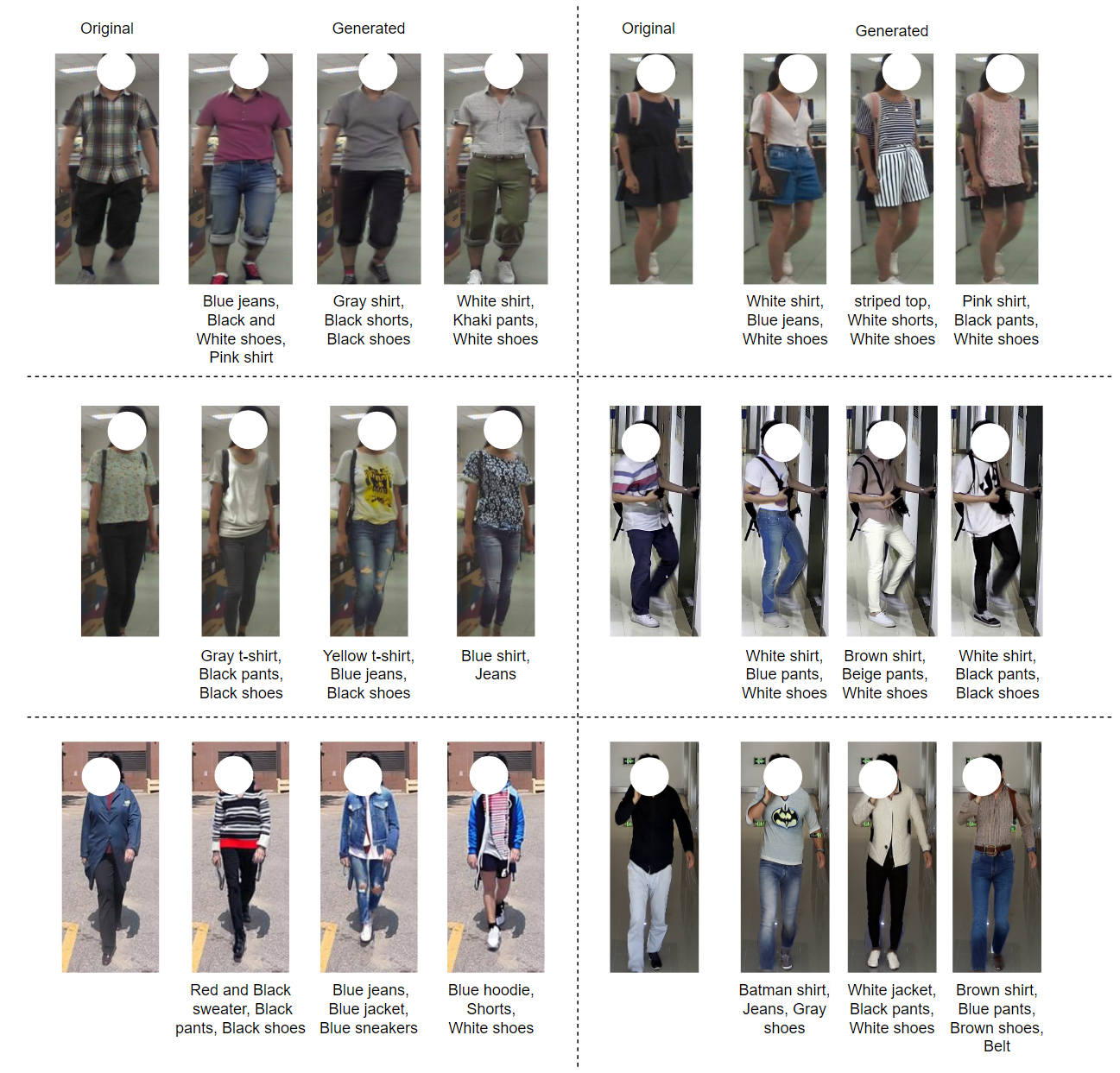}
    \caption{Qualitative examples of our generated data for PRCC (row 1), LTCC (row 2) and CCVID (row 3) datasets. For each original image, we show three inpainted versions. The prompts used to generate the inpainted samples are placed under the corresponding images. These samples depict high-quality, diverse generated data that is prompt-aligned.}
    \label{fig:qualitative-gen-supp}
\end{figure*}

Figure~\ref{fig:qualitative-gen-supp} showcases additional qualitative examples of our generated data spanning three datasets: PRCC, LTCC, and CCVID. These examples illustrate how our inpainted images respect the provided prompts, while also displaying the realism and diversity in the generated clothing. For example, in the top-left sample of Figure~\ref{fig:qualitative-gen-supp}, we see that the diffusion inpainting model properly generates all the different combinations of pants and shirts described in the prompts. In the top-right example, the inpainting model even completes slightly more difficult tasks, such as replacing a dress with two separate clothing items (blouse and shorts), while preserving the realism of the synthesized image. Leveraging diffusion to generate additional clothes-changed images is paramount for enriching training data, as DLCR is equipped to controllably and accurately increase the clothing diversity of any given dataset.

\label{add-qual-results}



\section{Improving Existing Methods using DLCR (cont.)}
\label{add-on-stage2}
In Table $3$ of the main paper, we showed how training any Re-ID model simply only using generated data from stage 1 of DLCR still yields large improvements. To fully exhibit the benefits of DLCR, we provide additional results in Table \ref{tab:stage2-addon}, where we apply both stages of DLCR to these models. When only using stage 1 of DLCR to train various Re-ID models, large improvements of $\approx7\%-28\%$ are observed across many models (middle rows of Table \ref{tab:stage2-addon}). With the introduction of progressive learning and prediction refinement in stage 2 of DLCR, top-1 accuracy on standard Re-ID models further increases by roughly $2-3\%$ for a cumulative increase of nearly $\approx10\%-30\%$ over the baseline  (last rows of Table \ref{tab:stage2-addon}). Similarly, top-1 accuracy on CC-ReID models improves by $\approx 1-4\%$ when adding stage 2 of DLCR, with the larger improvement possibly coming from the explicit clothes-invariance already instilled in these models. Thus, while stage 1 of DLCR can be applied to any model for significant performance gains, we further exhibit that using both stages yields the best results across many standard and CC-ReID models.

\begin{table}[ht!]
    \centering
    \small
     \caption{Results on PRCC when using both stages of DLCR on various standard Re-ID and CC-ReID models. Adding stage 2 results in better mAP and top-1 accuracy values on every model, with the green numbers in parentheses signifying cumulative improvement over the original baseline model. $\dagger$ denotes reproduced results using open-source code.}
     
    \begin{tabular}{c|c|c}
         \Xhline{3\arrayrulewidth}
         \multicolumn{3}{c}{Standard Re-ID Models}\\
         \Xhline{3\arrayrulewidth}
          Model & Top-1 & mAP \\
          \hline
         PCB \cite{Sun-ECCV-2018}  & 41.8 & 38.7\\
         PCB + DLCR (Stage 1) & 53.3 \textbf{(\textcolor{Green}{+11.5})} & 50.7 \textbf{(\textcolor{Green}{+12.0})}\\
         PCB + DLCR (Stage 1 + 2)  & \textbf{56.5 (\textcolor{Green}{+14.7})} & \textbf{51.0 (\textcolor{Green}{+12.3})}\\
         
         \hline
         MGN \cite{Wang-ICM-2018}  & 33.8 & 35.9\\
         MGN + DLCR (Stage 1)  & 62.5 \textbf{(\textcolor{Green}{+28.7})} & 57.6 \textbf{(\textcolor{Green}{+21.7})} \\ 
         MGN + DLCR (Stage 1 + 2)  & \textbf{64.8 (\textcolor{Green}{+31.0})} & \textbf{58.0 (\textcolor{Green}{+22.1})}\\
         
         \hline
         HPM \cite{Fu-AAAI-2018} & 40.4 & 37.2 \\
         HPM + DLCR (Stage 1)  & 56.0 \textbf{(\textcolor{Green}{+15.6})} & 50.9 \textbf{(\textcolor{Green}{+13.7})}\\
         HPM + DLCR (Stage 1 + 2)  & \textbf{57.5 (\textcolor{Green}{+17.1})} & \textbf{51.2 (\textcolor{Green}{+14.0})}\\
         
        \Xhline{3\arrayrulewidth}
        \multicolumn{3}{c}{CC-ReID Models}\\ 
        \Xhline{3\arrayrulewidth}
        
         Model & Top-1 & mAP  \\
         \hline
         CAL $\dagger$ \cite{Gu-CVPR-2022} & 55.2 & 55.8 \\
         CAL + DLCR (Stage 1)  & 62.9 \textbf{(\textcolor{Green}{+7.7})} & 60.9 \textbf{(\textcolor{Green}{+5.1})}\\
         CAL + DLCR (Stage 1 + 2)  & \textbf{66.5 (\textcolor{Green}{+11.3})} & \textbf{63.0 (\textcolor{Green}{+7.2})}\\
         
         \hline
         AIM $\dagger$ \cite{AIM} & 55.7 & 56.3 \\
         AIM + DLCR (Stage 1)  & 60.2 \textbf{(\textcolor{Green}{+4.5})} & 59.0 \textbf{(\textcolor{Green}{+2.7})}\\
         AIM + DLCR (Stage 1 + 2)  & \textbf{61.9 (\textcolor{Green}{+6.2})} & \textbf{60.5 (\textcolor{Green}{+4.2})}\\
         
         \hline
         GEFF \cite{Arkushin-WACV-2024} & 83.6 & 64.0 \\
         GEFF + DLCR (Stage 1)  & 84.6 \textbf{(\textcolor{Green}{+1.0})} & 66.0 \textbf{(\textcolor{Green}{+2.0})}\\
         GEFF + DLCR (Stage 1 + 2) &\textbf{85.8 (\textcolor{Green}{+2.2})} & \textbf{66.2 (\textcolor{Green}{+2.2})}\\ 
         \Xhline{3\arrayrulewidth}
    \end{tabular}
    \label{tab:stage2-addon}
\end{table}

\section{Visualizing DLCR's Improvements}
\label{visualizations}

\subsection{Qualitative Retrieval Examples}
\begin{figure*}[ht!]
    \includegraphics[width=\linewidth]{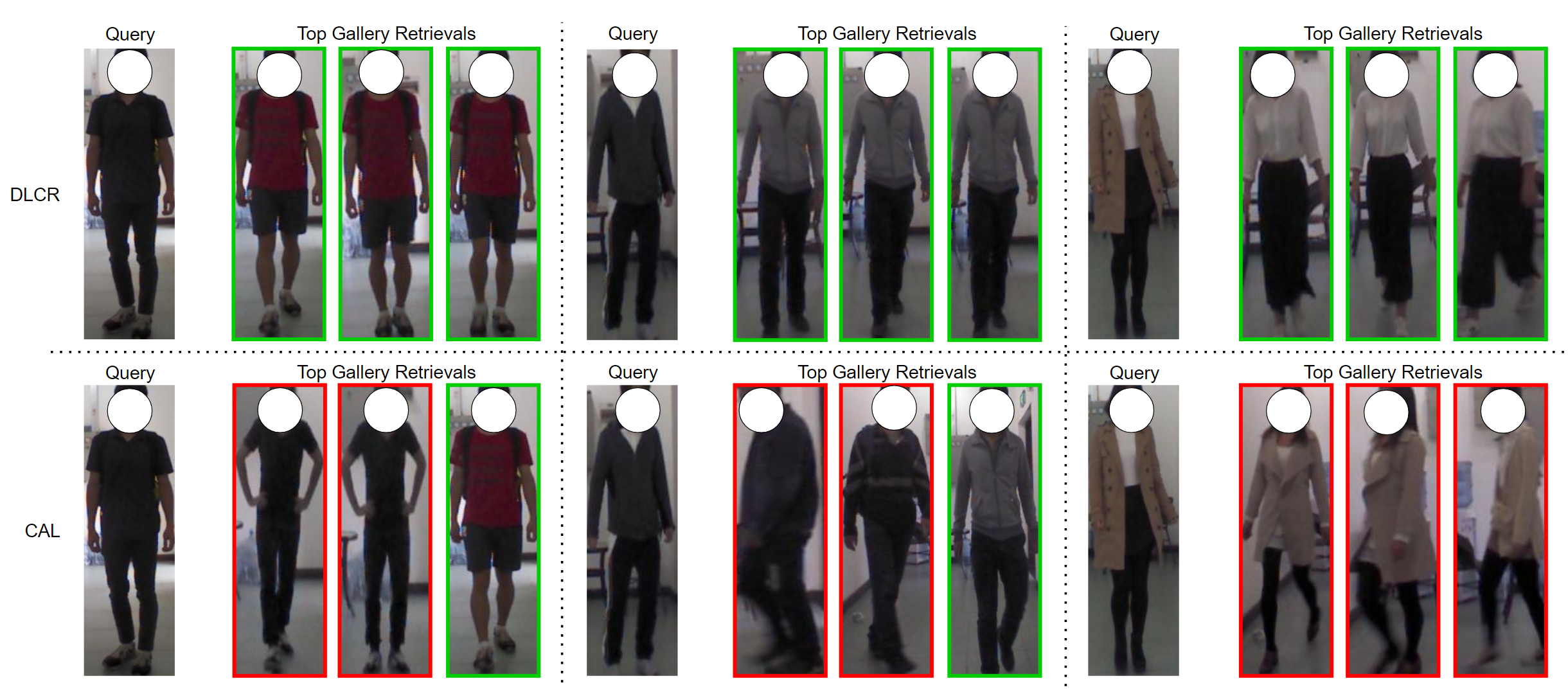}
    \caption{Qualitative retrievals of CAL+DLCR versus baseline CAL. For a given query image, the top-3 retrieved images from the gallery are shown, with correct and incorrect retrievals outlined in green and red, respectively. Despite clothing changes between the query and gallery images, CAL+DLCR retrieves the correct subject regardless of appearance. However, CAL still favors clothing items during retrieval, often retrieving incorrect subjects from the gallery that share similar clothing items to the query. This shows that discriminative approaches to clothing-invariance, such as CAL, can still be further improved using generative methods.}
    \label{qualitative-retrievals}
\end{figure*}

One way to visualize DLCR's improvement over CAL is shown in Figure \ref{qualitative-retrievals}, where we visualize the query-gallery retrievals for both models during evaluation. In the top half of the figure, DLCR correctly matches a query image with multiple gallery images of the same subject, regardless of the change in the subject's clothing. In the bottom half, we show how CAL fails on the same exact samples by erroneously retrieving images from the gallery of different subjects wearing similar clothing items to the query image. Despite the fact that CAL is explicitly designed for clothes-invariance, there still appears to be some bias towards clothing during evaluation. As we mention in the main paper, solely discriminative approaches to CC-ReID are not currently sufficient and leave significant room for improvement, such as utilizing generative approaches like DLCR. For example, one explanation for CAL's limitation could be the limited number of clothes-changes in the training data which prevents the full use of CAL's clothes-invariant learning strategy. Hence, DLCR better equips CC-ReID models to learn clothes-agnostic person features through the use of its generated data (stage 1) and training/testing strategies (stage 2).

\subsection{t-SNE Feature Plots}
\begin{figure*}[ht!]
    \centering
    \resizebox{\linewidth}{!}{
    \includegraphics[width=0.9\linewidth]{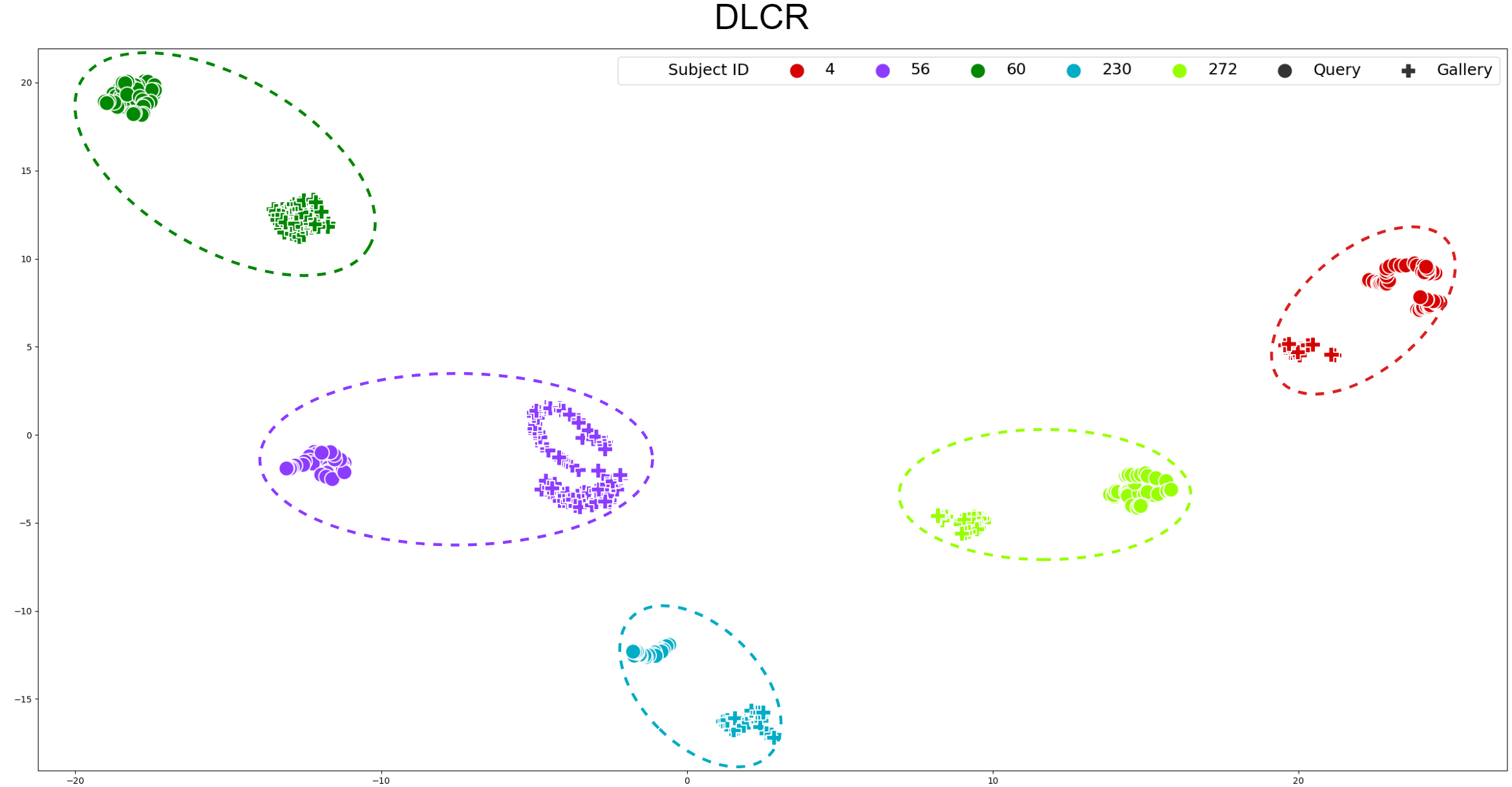}
    \includegraphics[width=0.9\linewidth]{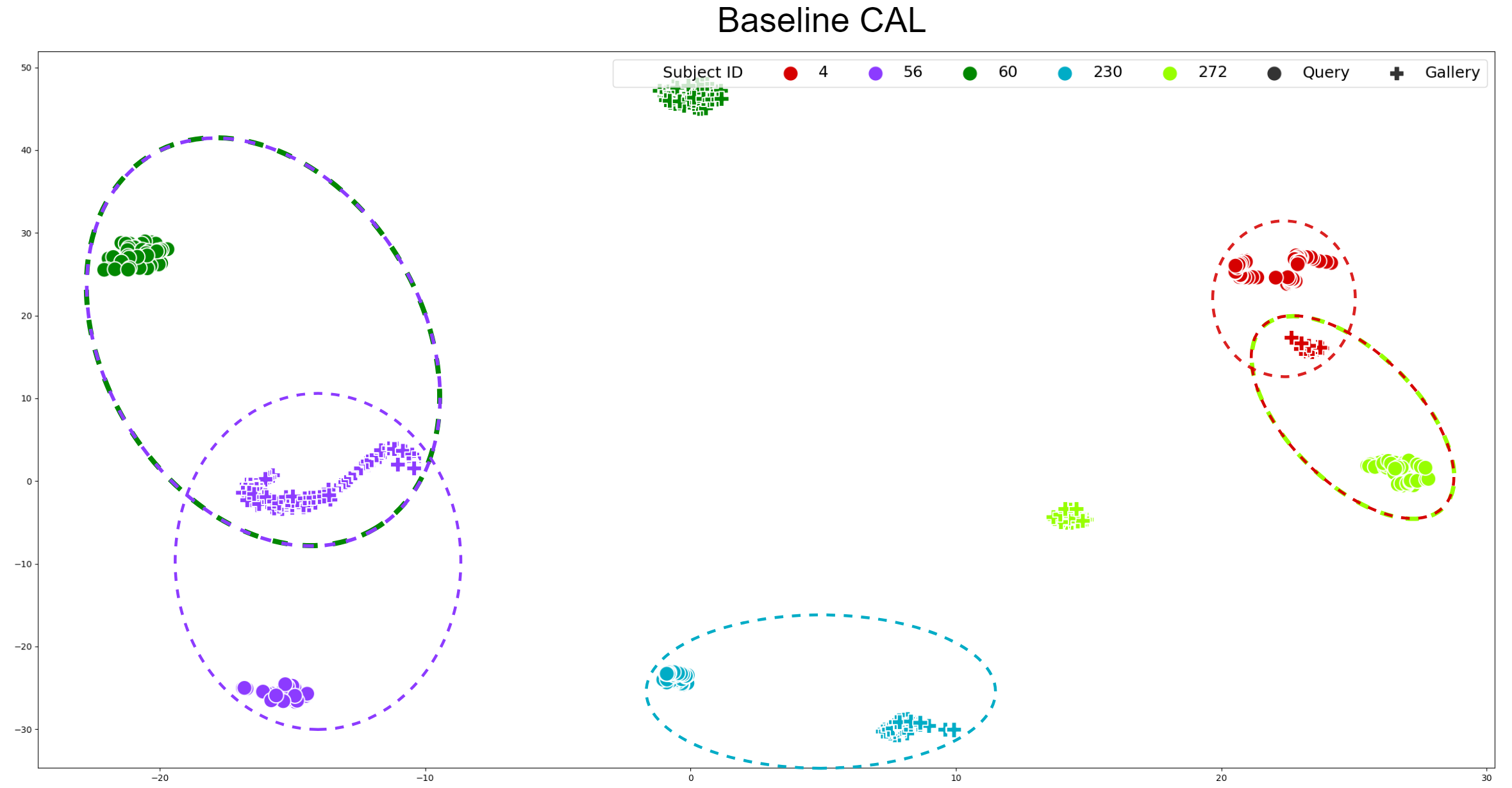}
    }
    \caption{t-SNE visualizations of the query and gallery features produced by CAL+DLCR and CAL for $5$ randomly selected test subjects in the PRCC dataset. With a baseline CAL model (right), the query feature cluster for Subject $272$ are erroneously closer to the gallery feature cluster for Subject $4$, with the same issue between Subject $60$ and $56$. The incorrect clustering behaviors are marked with multi-colored boundaries. In contrast, the gallery and query feature clusters produced by CAL+DLCR (left) for the same subjects correctly cluster together, exhibiting DLCR's direct impact in learning better discriminative features and improving top-1 accuracy.}
    \label{fig:tsne-comps}
\end{figure*}

As an additional way to visualize how DLCR improves top-1 accuracy during retrieval, Figure \ref{fig:tsne-comps} provides t-SNE plots to compare learned person features between a baseline CAL model and DLCR. As described in Sec. $4$ of the main paper, a query image is paired with a gallery image during evaluation by retrieving the gallery image with the most similar person features to the query image. An oracle model would produce identical person features for a query and gallery image if the same subject is in the image, regardless of clothing, background, occlusions, body pose, etc. In the left plot in Figure \ref{fig:tsne-comps}, we show the resulting query and gallery person features from DLCR for five randomly selected subjects in the PRCC testing set. For each query and gallery image of a particular subject, DLCR correctly produces person features that cluster in both an inter- and intra-class fashion. Not only does each gallery feature of a particular subject cluster with other gallery features of the same subject, but the same clustering behavior occurs between each query feature of a particular subject as well. More importantly, the smallest distance between a query feature cluster and gallery feature cluster produced by DLCR share the same subject ID, indicating correct test-time retrievals. However, this behavior is not seen for the same exact subjects and samples when using a baseline CAL model, as shown in the right plot of Figure \ref{fig:tsne-comps}. CAL does not produce similar person features between query and gallery images of the same subject, with the erroneous query-gallery clustering examples explicitly highlighted with multi-colored boundaries. For example, the query feature cluster for Subject $272$ is closer in distance to the gallery feature cluster of Subject $4$, decreasing top-1 performance since the wrong gallery image would be retrieved.  

\subsection{Activation Maps}
\begin{figure}[h!]
    \centering
    \includegraphics[width=0.95\linewidth]{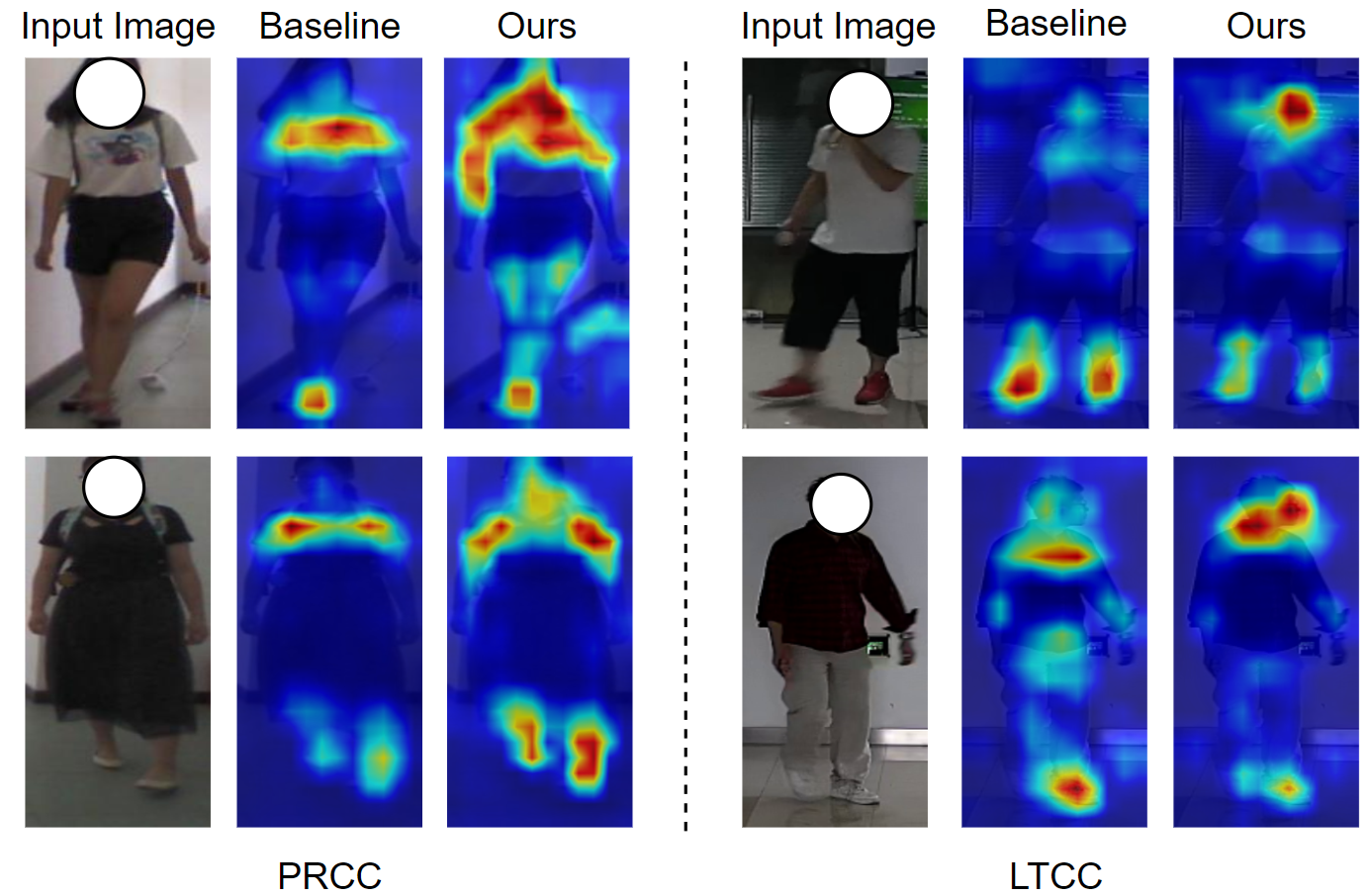}

    \caption{Feature map comparison between DLCR and a baseline CAL model. DLCR enforces better robustness against clothing variations in CC-ReID.
    }
    \label{fig:activation-maps}
\end{figure}

In Figure~\ref{fig:activation-maps}, we compare the feature maps of CAL+DLCR with a baseline CAL model on the PRCC and LTCC datasets. Notably, CAL+DLCR exhibits a stronger focus on identity-related features. For instance, the DLCR feature map in the LTCC examples prioritizes the subject's face over the footwear when making a prediction. Furthermore, DLCR retains the ability to leverage person-specific features that are within the clothing region (\eg body shape) despite its invariance towards clothing, as seen in the PRCC examples.

\section{Optional: Discriminator-guided diffusion}
\label{sec:disc-guidance}
Considering our downstream task of Re-ID training, it is important to ensure that $G_{train}$ closely resembles the distribution from which $D_{train}$ originated. The generated set $G_{train}$ is obtained using a pretrained diffusion model, but it has been shown that without additional fine-tuning, there can be a moderate gap between the diffusion model's generated data and the real data distribution \cite{Dongjun-ICML-2023}. At the same time, fine-tuning a diffusion model can become computationally expensive and may lead to undesired results, like overfitting \cite{Dongjun-ICML-2023, Nichol-ICML-2021}. To avoid these problems, similar to \cite{Dongjun-ICML-2023}, we investigated using a discriminator $d_\phi$ to guide the pretrained diffusion model to generate data that is better aligned with the training data distribution. Following~\cite{Dongjun-ICML-2023}, we train the discriminator $d_\phi$ to minimize the domain gap with respect to the noisy examples at different timesteps from the real ($D_{train}$) and generated ($G_{train}$) data sets. Consequently, the employed training objective of $d_\phi$ (\ie~$\mathcal{L}_d$) is to distinguish between the real and generated examples:
\begin{equation}
    \label{loss_disc}
    \mathcal{L}_d = -\mathbb{E}\left[\log{d_\phi(x_t, t)} + \log{\left(1 - d_\phi(\hat{x}_t, t)\right)}\right],
\end{equation}
where $t \sim \mathcal{U}([0, T])$, $x_t \sim q(x_t|x), x \sim \mathcal{U}(D_{train})$ and $\hat{x}_t \sim q(\hat{x}_t|\hat{x}), \hat{x} \sim \mathcal{U}(G_{train})$.

We use the trained discriminator 
to design a score function that guides the generative process 
to synthesize samples that are 
highly likely to be classified as real by the discriminator. The respective score function is the following:
\begin{equation}
    \label{score_disc_guidance}
    h_\phi = \nabla_{x_t} \log{\frac{d_\phi(x_t, t)}{1 - d_\phi(x_t, t)}}.
\end{equation}
We incorporate this score function $h_\phi$ 
into 
noise estimation 
as follows (see Sec. \ref{disc-derivation} for the full derivation):
\begin{equation}
    \label{noise_score}
    \epsilon_\phi(x_t, t) = -\sigma_t \cdot h_\phi,
\end{equation}
where $\sigma_t=\sqrt{1-\bar{\alpha}_t}$ is the standard deviation of $q(x_t|x_0)$. Then, we 
guide 
the reverse denoising process of Stable Diffusion by 
adding the noise estimation from our discriminator, 
as follows:
\begin{equation}
    \label{final_noise_estimation}
    \epsilon^{\phi}_{\theta}(x_t, t) = \epsilon_\theta(x_t, t) + w\cdot \epsilon_\phi(x_t, t),
\end{equation}
where $w$ denotes the weight 
of discriminator guidance.

Finally, we generate an improved version of $G_{train}$ 
by modifying the mean $\boldsymbol{\mu}_\theta(x_t, t)$ of the reverse denoising process 
(Eq.~$(3)$) to include the discriminator-guided noise estimation $\epsilon_\phi(x_t, t)$:
\begin{equation}
    \label{mean_reverse_process}
    \boldsymbol{\mu}^\phi_\theta(x_t, t) = \frac{1}{\sqrt{\alpha_t}}\left(x_t - \frac{1-\alpha_t}{\sqrt{1-\Bar{\alpha}_{t}}}\epsilon^{\phi}_{\theta}(x_t,t)\right).
\end{equation}

We found that in some cases, using discriminator-guided diffusion improved results on PRCC and CCVID by $1-2\%$, however the process is optional since it requires some additional training of a discriminator and re-generation of the data (following the process proposed in \cite{Dongjun-ICML-2023}). In certain cases where data using a pretrained diffusion model is not generating sufficiently in-domain data, we included this section as one potential solution as well as a strong area for future work.

\subsection{Details Regarding Discriminator Guidance}
\label{disc-derivation}
In this section, we will present the derivations for Eq.~$(6)$ and Eq.~$(7)$, while also providing more intuitive explanations for them. 
Before describing these details, we remind the reader that in the continuous formulation \cite{Song-ICLR-2021} of diffusion models, the forward process is described by a stochastic differential equation (SDE):
\begin{equation}
    \label{forward_SDE}
    \partial x_t = f(x_t, t)\partial t + g(x_t)\partial w,
\end{equation}
and the reverse process is also a diffusion process \cite{Song-ICLR-2021, Anderson-SPA-1982}, given by the following SDE:
\begin{equation}
    \label{reverse_SDE}
    \partial x_t = [f(x_t, t) - g^2(x_t)\nabla_x \log{q(x_t)}]\partial t + g(x_t)\partial\bar{w},
\end{equation}
where $f$ is called the drift coefficient, $g$ denotes the diffusion coefficient and $\nabla_x \log{q(x_t)}$ is called the score function. This score function is what is estimated by diffusion models in order to solve the reverse process.

Regarding discriminator guidance \cite{Dongjun-ICML-2023}, we emphasize that its objective is to address the situations when the diffusion model converges to a local optimum, thereby failing to provide the most accurate noise or score estimations. In a formal sense, when dealing with these cases, it becomes necessary to correct the marginal distribution $p_\theta(x_t)$ of the forward process (Eq.~\eqref{forward_SDE}), which originates with samples drawn from $p_\theta(x_0)$, in order to match the marginal distribution $q(x_t)$ of the forward process initiated with samples from $q(x_0)$. Dongjun \etal~\cite{Dongjun-ICML-2023} introduce this correction term as an additional score function that depends on a discriminator, and the term is used in the reverse process at each denoising step. We derive this term starting with the following simple observation:
\begin{equation}
    q(x_t) = p_\theta(x_t) \cdot \frac{q(x_t)}{p_\theta(x_t)},
\end{equation}
and then, if we apply the logarithm and the gradient, we get:
\begin{equation}
    \label{ideal_score_function}
    \nabla_x\log{q_(x_t)} = \nabla_x\log{p_\theta(x_t)} + \nabla_x\log{\frac{q(x_t)}{p_\theta(x_t)}}.
\end{equation}
Eq.~\eqref{ideal_score_function} 
implies that if we want to obtain the optimal score function ($\nabla_x \log{q(x_t)}$ - the one required to solve Eq.~\eqref{reverse_SDE}), we can 
correct the model score estimation $\nabla_x\log{p_\theta(x_t)}$ using the log-gradient of the rate $\frac{q(x_t)}{p_\theta(x_t)}$. However, we cannot compute this rate in practice, thereby Dongjun \etal~\cite{Dongjun-ICML-2023} propose to estimate it via a discriminator. More precisely, 
a discriminator, $d_\phi(x_t, t)$, is trained to distinguish between real and generated samples. After the training is completed, $d_\phi(x_t, t)$ will return the probability for the sample $x_t$ of being a real example at every timestep. Thus, it is an estimation for $q(x_t)$ and, in a similar manner, $1 - d_\phi(x_t, t)$ will approximate $p_\theta(x_t)$. Therefore, we use:
\begin{equation}
     h_\phi = \nabla_{x_t} \log{\frac{d_\phi(x_t, t)}{1 - d_\phi(x_t, t)}}
\end{equation}
 as a correction term in the reverse process, which is the same $h_\phi$ as defined in Eq.~$(6)$. 

Moving further, Eq.~$(7)$ denotes the relation between the noise estimation and the corresponding score function. We will derive a more general form of this equation by exploiting the reparameterization trick for a Gaussian distribution and Tweedie's formula \cite{Robbins-SMSP-1956}. 

Given an arbitrary Gaussian distribution $\mathcal{N}(x;\mu,\sigma^2\mathbf{I})$ and its corresponding density function $p(x)$. The reparameterization trick applied for distribution is the following:
\begin{equation}
    \label{reparameterization_trick}
    \begin{split}
                x &= \mu + \sigma\cdot\epsilon 
                \iff \mu = x - \sigma\cdot\epsilon, \epsilon \sim \mathcal{N}(0, \mathbf{I}).
    \end{split}
\end{equation}

In statistical literature, Tweedie's formula generally shows how to express the mean of an arbitrary Gaussian distribution ($\mu$) given its samples ($x$) and the score function ($\nabla_x\log{p(x)}$). Specifically, we apply Tweedie's formula 
on the previous Gaussian distribution and we obtain the following result:
\begin{equation}
    \label{tweedie}
    \begin{split}
        \mu = x + \sigma^2 \nabla_x \log{p(x)}.
    \end{split}
\end{equation}
If we combine the results from Eq.~\eqref{reparameterization_trick} and Eq.~\eqref{tweedie}, then we obtain the following:
\begin{equation}
    \label{}
    \begin{split}
        \epsilon = -\sigma\nabla_x \log{p(x)}, 
    \end{split}
\end{equation}
and we can apply this result for the discriminator score function denoted by $h_\phi$ and $\sigma = \sqrt{1-\Bar{\alpha_t}}$ to obtain Eq.~$(7)$:
\begin{equation}
    \label{}
    \begin{split}
        \epsilon_\phi(x_t, t) = -\sqrt{1-\Bar{\alpha}_t}h_\phi. 
    \end{split}
\end{equation}

\section{Utility of LLaVA and LLaMA}
\label{sec:llm-motif}
In this section, we 
provide reasoning behind utilizing LLaVA and LLaMA to extract clothing descriptions of the clothing IDs present in a dataset.

In Table~\ref{tab:llm-importance}, we provide the results obtained for three cases on the PRCC dataset. In the first row, we ask only LLaMA to give us random sets of clothing items for each specific body part (top, bottom, footwear) for clothes inpainting. In the second row, we only use LLaVA to create clothing prompts from a single image of a given clothing ID (no use of LLaMA summarization). We compare these two ablations to DLCR's main results in the third row, where both LLaVA and LLaMA are used to construct prompts for clothes inpainting. Table \ref{tab:llm-importance} highlights that it is beneficial to generate data from clothes that are already (or close to) present in the dataset. This aspect is intuitive because during training, if two subjects are wearing relatively identical clothing items, a CC-ReID model cannot exploit the clothing information to classify a subject. Thus, the model must rely on ID-specific features to differentiate between the subjects.
Moreover, we illustrate the robustness of our method to different LLMs in Table \ref{tab:diff_vlm},
where we compare LLaVA with InstructBLIP \cite{dai2305instructblip} for extracting textual clothing descriptions. We observe very similar results, implying that the selection of LLMs has a minimal impact on DLCR's performance.
Overall, our use of LLaVA and LLaMA is beneficial for CC-ReID performance, as seen in the $\approx2\%$ increase in both top-1 accuracy and mAP.

\begin{table}[ht!]
    \centering
    \caption{Results on the PRCC dataset when generating data with and without extracted text prompts from LLaVA and LLaMA. Cumulative improvements over the baseline are shown in green.}
    \resizebox{\columnwidth}{!}{
    \begin{tabular}{c|c|c|c}
    \hline
    \multirow{2}*{LLaVA} & \multirow{2}*{LLaMA} & \multicolumn{2}{c}{PRCC} \\
    \cline{3-4}
    & & Top-1 & mAP \\
    \Xhline{3\arrayrulewidth}

    \textcolor{red}{\xmark} & \textcolor{Green}{\cmark} (random clothing prompts) & 60.7 & 58.9 \\
    \textcolor{Green}{\cmark} & \textcolor{red}{\xmark} & {61.6 (\textcolor{Green}{+0.9})}  & {60.1 (\textcolor{Green}{+1.2})} \\
     \textcolor{Green}{\cmark} & \textcolor{Green}{\cmark} & \textbf{62.9 (\textcolor{Green}{+2.2})}  & \textbf{60.9 (\textcolor{Green}{+2.0})} \\
    \hline
    \end{tabular}
    }
    
    \label{tab:llm-importance}
\end{table}

\begin{table}[ht!]
    \centering
        \caption{Stage $1$ DLCR results on PRCC with different VLMs.}
    \begin{tabular}{c|c|c}
        Visual Language Model (VLM) & Top-1 & mAP  \\
        \Xhline{3\arrayrulewidth}
        LLaVA & 62.9 & 60.9 \\
        InstructBlip & 62.9 & 61.3
    \end{tabular}
    \label{tab:diff_vlm}
\end{table}






\section{Progressive Learning Intuition}
\label{sec:prog-learning}
\begin{figure}[ht!]
    \includegraphics[width=\linewidth]{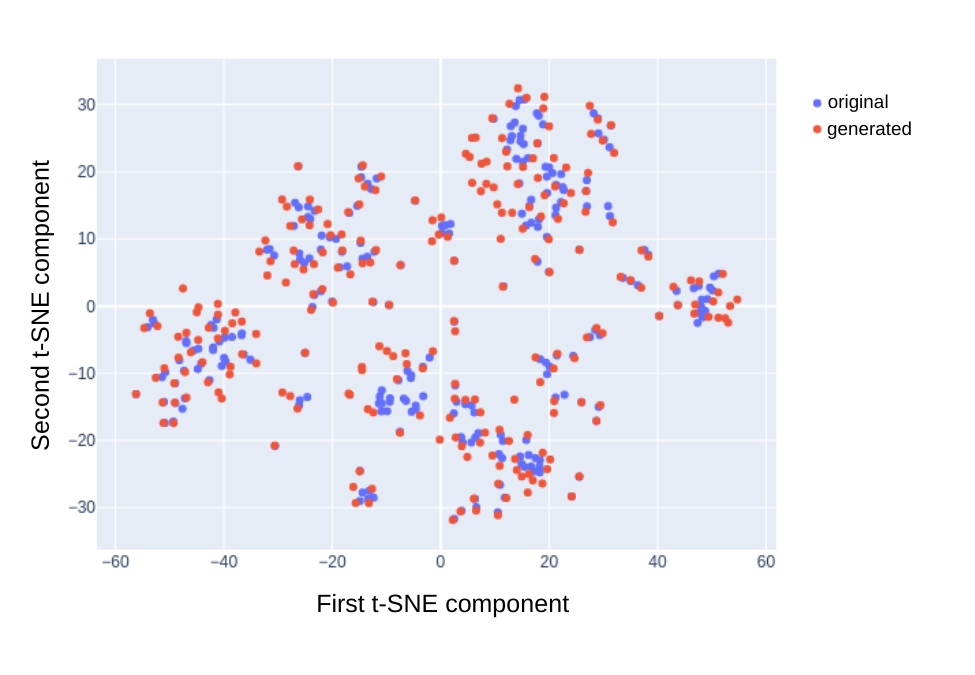}
    \caption{t-SNE plot of our generated data versus the original data for three random subjects in the PRCC dataset. As our generated data introduces a larger variance in the training distribution, our progressive learning strategy is effective in optimizing CC-ReID performance.}
    \label{generated-data-vs-real-data}
\end{figure}

As discussed in Sec. \ref{sec:prog-learning}, one limitation of generating so much additional data would be the impact on training time. The choice to inject the generated data at the batch level was largely driven by this point, as shown in Table \ref{tab:prog-train-time}. On the other hand, Figure~\ref{generated-data-vs-real-data}, provides a t-SNE plot to visualize and compare the principal components between our generated images and the original images for three random subjects from the PRCC dataset. It is easy to notice that the inclusion of generated samples leads to a larger variance in the data, thus making the task of differentiating between users more challenging. To alleviate this issue at the early stages of training, we use a smaller number of generated samples and gradually incorporate additional generated samples as training progresses. This strategy allows the model to more effectively adapt to the increasingly diverse distribution of the generated data, as illustrated by the performance shown in Table \ref{tab:ablations}, row $5$.

\begin{table}[ht!]
    \caption{Training times with and without progressive learning when using DLCR generated data. With identical experimental setups, progressive learning not only reduces training time, but also provides some performance boost  (see Table \ref{tab:ablations}).}
    
    \centering
    \resizebox{\linewidth}{!}{
    \begin{tabular}{c|c|c|}
         Progressive Learning & PRCC Training Epoch Time (seconds) $\downarrow$ & Best Top-1 $\uparrow$ \\
        \Xhline{3\arrayrulewidth}
         \textcolor{red}{\xmark} & 691s & 64.0 \\
         \textcolor{Green}{\cmark} & 65s & 65.0 \\
    \label{tab:prog-train-time}
    \end{tabular}
    }
\end{table}

\section{Ablation on Number of Generated Images ($K$)}

\begin{figure}[ht!]
    \centering
    \includegraphics[width=\linewidth]{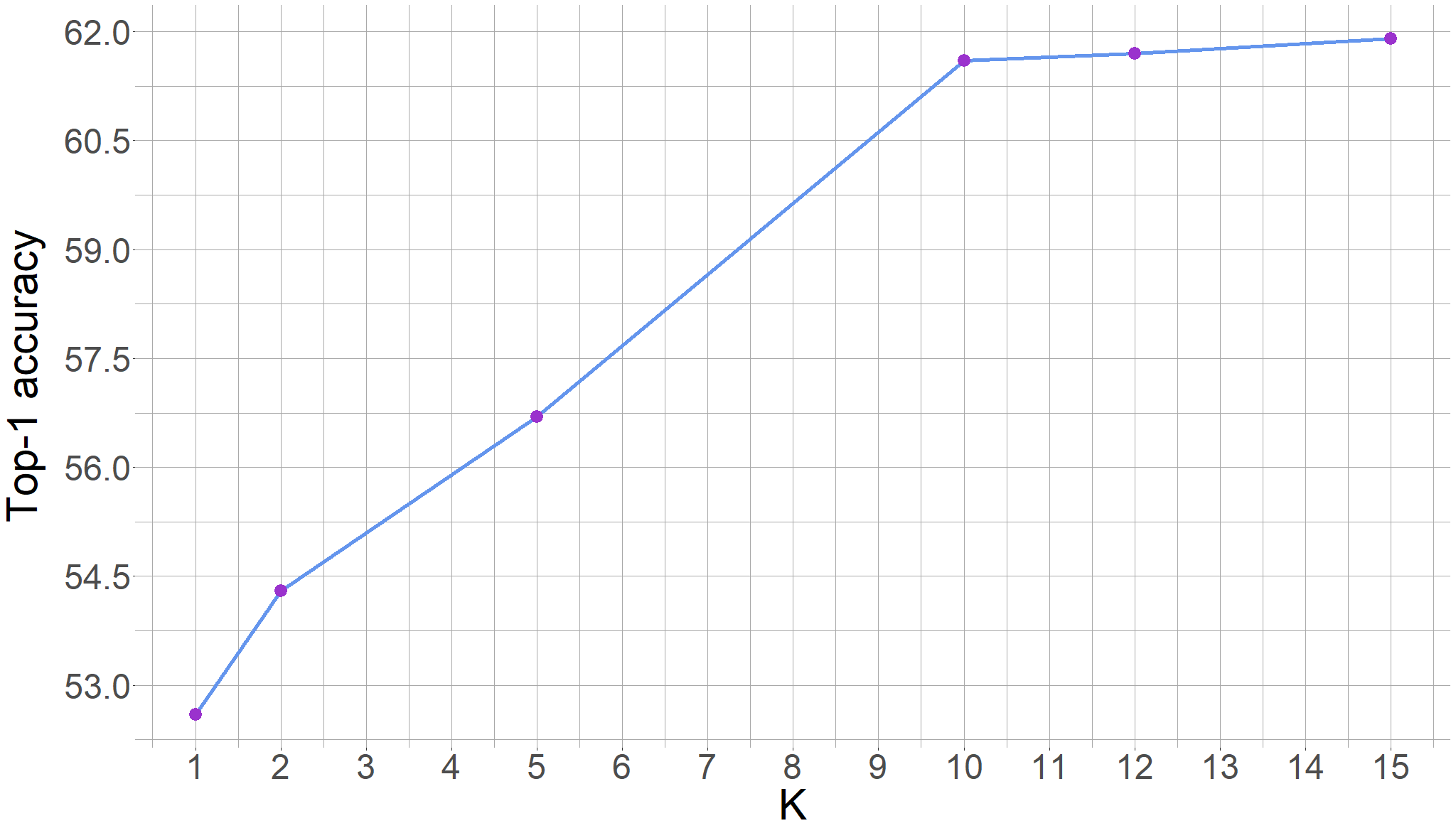}
    \caption{ Top-1 accuracy rates achieved for various values of $K$ (number of inpainted images per original image) by DLCR on PRCC dataset. The accuracy improvements for $K \geq 10$ are marginal. Therefore, we standardized the value of $K$ to 10 for all our experiments.}
    \label{fig:ablation-k}
\end{figure}

As detailed in the implementation details of the main paper, we set the number of inpainted versions for each original training image $K=10$. In Figure~\ref{fig:ablation-k}, we perform an ablation on the values of $K$ and provide the resulting top-1 accuracies on the PRCC dataset. Notably, the improvements observed when increasing $K$ beyond 10 are minimal, hence explaining why we do not simply generate more images to increase the size of our contributed data. We set $K=10$ as it strikes a favorable balance between the performance gain facilitated by the additional generated data, and the time consumed by the generation process. However, the DLCR generation process is open-source and can be used by others to generate more data for cases such as large-scale CC-ReID pretraining. Note that the results presented in Figure~\ref{fig:ablation-k} are obtained by simply concatenating the generated data with the initial training set of PRCC, \ie~we do not perform progressive learning, discriminator guidance or prediction refinement in this study.

\label{sec:clothing-ablation}

\section{Time and Space Complexity of DLCR}
\label{sec:time}
We provide explicit time and space complexities for each component of DLCR in Table \ref{tab:time-space-comp}. Specifically, we report the throughput of each of our components, measured by how many images each component can process per second, as well as the memory each component takes on the GPU.  Despite our use of LLMs and diffusion, DLCR as a whole is still fairly computationally inexpensive since we only use these large models in an inference setting. During the generation of CC-ReID training data via diffusion inpainting, which is an offline process (Stage 1 of DLCR), we use $50$ denoising steps at a resolution of $768 \times 256$. Then we apply stage $2$ and train the CC-ReID model with the combination of original and generated data. 
During inference, for query inpainting in our prediction refinement strategy, we reduce the denoising steps from $50$ to $10$ and divide the image resolution by half, which results in a nearly $\sim3\times$ speedup while maintaining similar performance. In summary, stage 1 of DLCR can be fully implemented on a single NVIDIA A100 80GB GPU (or a A6000 48GB GPU with some tricks to fit the LLMs). 

\begin{table}[ht!]
    \centering
    \caption{Analysis of time and space complexity for each component in DLCR. Throughput is measured in images/sec, with all experiments run on a single NVIDIA A100 80GB GPU.}
    \resizebox{\linewidth}{!}{
    \begin{tabular}{c|c|c}
         DLCR Component & Throughput (img/sec) $\mathbf{\uparrow}$ & Memory $\mathbf{\downarrow}$  \\
         \Xhline{3\arrayrulewidth}
         ID-Preserving Mask Extraction & 61.6 & 1GB \\
         Clothes Description Extraction & 4.0 & 56GB \\
         Training Inpainting & 0.89 & 4GB \\
         Query Inpainting & 3.33 & 2.6GB \\
         Prediction Refinement (Algorithm $1$) & 642.6 & $<$ 1GB\\
        \hline
    \end{tabular}
    }
    \label{tab:time-space-comp}
\end{table}

\section{Limitations and Future Work}
\label{sec:limitations}
\noindent\textbf{Limitations:} One limitation of DLCR is low-quality generated images on low-resolution images. On low-resolution or small-scale images, not only can the ID-preserving mask be inaccurate, but the diffusion model itself struggles to properly inpaint the clothing regions correctly. Some datasets mentioned in this paper have these types of images, and the generated data on these small corner case images can be low quality and incorrect/not prompt-aligned. Due to our use of a pretrained Stable Diffusion model, one possible solution is to fine-tune Stable Diffusion on low-resolution images for better performance. Diffusion models have also shown strong promise in image super-resolution, which could be included in the DLCR generation pipeline to deal with low-resolution images. Another limitation could be the diminishing returns of additional data (Fig. \ref{fig:ablation-k}). While DLCR will most likely show significant performance gains in data-scarce domains, finding a method to better leverage mass amounts of data while mitigating diminishing returns would heavily impact the positive effect DLCR has on downstream performance. 

\noindent\textbf{Future Work:} One possible direction of future work for DLCR could be investigating our test-time prediction strategy compared to other re-ranking methods. In the context of CC-ReID, we show that our prediction refinement strategy is compatible with other re-reranking methods such as GEFF \cite{Arkushin-WACV-2024}, but further experimentation could yield even higher SOTA results. Another avenue of future use is leveraging the DLCR generation method for other vision tasks that are compatible with localized image editing. For example, our method of generating data with preserved areas using a binary mask could apply to medical imaging for data augmentation. 
\end{document}